\documentclass{article}

\pdfoutput=1
\usepackage{arxiv}

\usepackage[utf8]{inputenc} % allow utf-8 input
\usepackage[T1]{fontenc}    % use 8-bit T1 fonts
\usepackage{hyperref}       % hyperlinks
\usepackage{url}            % simple URL typesetting
\usepackage{booktabs}       % professional-quality tables
\usepackage{amsfonts}       % blackboard math symbols
\usepackage{amssymb}
\usepackage{amsmath}
\usepackage{nicefrac}       % compact symbols for 1/2, etc.
\usepackage{microtype}      % microtypography
\usepackage{graphicx}
\usepackage{subfig}
\usepackage{multirow}
\usepackage{natbib}
\setcitestyle{authoryear,round}

\title{Mango Tree Net - A fully convolutional network for semantic segmentation and individual crown detection of mango trees}

\author{
  Vikas Agaradahalli Gurumurthy\thanks{Address all correspondence to: Vikas Agaradahalli Gurumurthy. This research was carried out by the author at the Department of Aerospace Engineering, Indian Institute of Science.} \\
  Independent Researcher\\
  Bengaluru, India \\
  \texttt{ag.vikas98@gmail.com} \\
  \And
  Ramesh Kestur \\
  Bangalore Institute of Technology\\
  Bengaluru, India \\
  \texttt{rkestur@gmail.com} \\
  \AND
  Omkar Narasipura \\
  Indian Institute of Science \\
  Bengaluru, India \\
  \texttt{omkar@iisc.ac.in} \\
}

\begin{document}
\maketitle

\begin{abstract}
This work presents a method for semantic segmentation of mango trees in high resolution aerial imagery, and, a novel method for individual crown detection of mango trees using segmentation output. Mango Tree Net, a fully convolutional neural network (FCN), is trained using supervised learning to perform semantic segmentation of mango trees in imagery acquired using an unmanned aerial vehicle (UAV). The proposed network is retrained to separate touching/overlapping tree crowns in segmentation output. Contour based connected object detection is performed on the segmentation output from retrained network. Bounding boxes are drawn on the original images using coordinates of connected objects to achieve individual crown detection. The training dataset consists of $8,824$ image patches of size $240\times240$. The approach is tested for performance on segmentation and individual crown detection tasks using test datasets containing $36$ and $4$ images respectively. The performance is analyzed using standard metrics precision, recall, f1-score and accuracy. Results obtained demonstrate the robustness of the proposed methods despite variations in factors such as scale, occlusion, lighting conditions and surrounding vegetation.  
\end{abstract}

% keywords can be removed

\section{Introduction}

Mango is a fruit of great economic importance in the tropical countries of South Asia. More than half of the world’s mango production comes from India alone with 2,309,000 acres of mango farms. The revenue of mango exports from India amounted to more than \$50 million between January to April 2017. Mangoes have a great demand both in fresh food and processed food industries. Increase in global population is expected to place a greater demand for mango. The major mango producer, India, has projected a production target of 36.92 million tonnes in the year 2030 \citep{mitra2016}. The growing demand for mango can be met by increasing acreage of cultivation and adopting new and innovative techniques to increase the yield. Information such as health, growth rate, flowering, flower to fruit conversion rate gathered at regular intervals during fruiting season facilitate effective decision making in mango cultivation. Besides providing adequate irrigation and fertilization, practices like pruning promote productivity. Capturing and converting sunlight into fruit biomass is a key process in fruit production. Pruning allows better light penetration and promotes reproductive phase over vegetative phase which results in increased yield \citep{gopu2014}. Manual assessment of tree crowns in a vast stretch of land is painstaking and often inaccurate. Moreover, manual data collection is labor intensive, costly and time consuming. Gathering data of vast expanses of mango farms with multi-temporal resolution manually is near impossible task. Remote sensing is the potential candidate to overcome these problems. The remote sensed images processed by computer vision algorithms provide meaningful and actionable information. The first step to automation of analysis of mango trees is their detection in remote sensed images. In this paper, we use a fully convolutional neural network based method for semantic segmentation of mango tree crowns. We further suggest a novel method for detection of individual mango tree crowns and obtaining their count.

Traditional satellite-based remote sensing has been used for vegetation cover analysis, geological studies, urban planning etc. \citep{feng2016} \citep{harald2016} \citep{weng2002}, where macro-level information is extracted using images with high spectral resolution and low spatial resolution. Micro-level information extraction for tasks such as tree species identification, requires images with high spatial resolution. Over the past few years, use of unmanned aerial vehicles (UAVs) for remote sensing – known as low altitude remote sensing (LARS) – has proliferated in agricultural research, crop and forest monitoring \citep{senthilnath2016} \citep{huang2018} \citep{baena2017}. Low altitude remote sensing allows for acquisition of high spatial resolution imagery at high temporal resolution, which is a cost-efficient alternative to satellite-based remote sensing. UAVs equipped with necessary payloads can be used in agriculture to create digital map of field, analyze crop health, find leaks in irrigation, find missing livestock etc \citep{reinecke2017}. UAVs find application in forest mapping, forest management planning, canopy height model creation and mapping forest gaps \citep{banu2016}. Analysis of trees in a plantation/orchard using imagery acquired through low altitude remote sensing is a viable and efficient approach.

Past research works have explored various traditional methods for detection, segmentation and delineation of tree crowns. Some of the well known methods include enhancement and thresholding, region growing, template matching, local maximum filtering and watershed segmentation. \citet{dralle1997} performed smoothing operation using gaussian-kernel and applied threshold to extract estimates of tree locations. To estimate positions of trees at ground level they used a displacement model incorporating the angle to the sun, the camera position and estimated tree heights. \citet{pollock1998} detected tree top using correlation between a single tree crown template from scene model – includes geometric-radiometric conditions during image acquisition and tree crown parameters – and image data. \citet{zhen2014} used marker controlled region growing for individual tree crown delineation. The tree tops detected using local maximal filtering were used as markers and six conditions based on homogeneity, crown area, and crown shape formed stop criteria for the region growing algorithm. \citet{huang_2018} applied local intensity clustering and morphological operations to smoothen the fine texture of canopy. They extracted individual tree crowns using marker-controlled watershed segmentation algorithm. Local maximum filtering was used to generate markers. \citet{baena2017} used object-based image analysis (OBIA) to identify and quantify tree species in very high resolution UAV images. \citet{kasper2018} used geographic object-based image analysis (GEOBIA) and eCognition developer software to delineate the tree crowns. 

\citet{morten2011} compared six traditional tree crown detection algorithms under varying forest conditions. The research found that none of the algorithms considered could analyze all the forest types - homogeneous, isolated and dense. The traditional methods are extremely sensitive to changes in scale, lighting conditions, plantation density and tree crown structure. The problem would exacerbate analyzing very high resolution UAV imagery. Hence, a more robust algorithm that is invariant to varying conditions is required. Recently developed state-of-the-art convolutional neural networks (CNN) could eliminate the shortcomings of traditional methods.

Since the advent of the convolutional neural networks (CNN), they have become benchmark in most computer vision problems. They have outperformed the traditional computer vision algorithms in the tasks of classification, object detection, semantic and instance segmentation. Several works have explored application of CNNs in the field of agriculture and forest management. \citet{carpentier2018} used CNN to classify the barks of different trees. \citet{guirado2017} compared CNN based methods with OBIA-methods. Their research proved CNN based methods to be exceedingly better than state-of-the-art OBIA-methods. \citet{kestur2018} used CNN based approach to detect and count mangoes in an orchard. \citet{li2017} used LeNet – a CNN based model – to detect the palm tree at the center of each patch extracted using a sliding window. They merged the multiple detection of same tree into one using an iterative method. A particular class of CNN known as fully convolutional neural networks (FCN) are widely used today in the semantic segmentation tasks. \citet{long2014} showed that CNN trained end-to-end, pixel-to-pixel can out-do existing methods in semantic segmentation. Their model achieved state-of-the-art performance in segmentation of PASCAL VOC, NYUDv2, SIFT Flow. \citet{kestur2018} adopted an approach to detect mangoes that involved an FCN to segment mangoes and detecting connected objects in segmentation output to detect mangoes.

Our research is aimed at developing algorithms for semantic segmentation of mango tree crowns and the detection of individual mango tree crowns by drawing bounding boxes around them. Semantic segmentation of the mango tree crowns is achieved using an FCN based approach. For detection of individual crowns, the approach adopted in this work is similar to that used by \citet{kestur2018}. The tree crowns are first segmented using the FCN based method. Connected objects are detected in the segmentation output and bounding boxes are drawn around those regions in the original image. The shortcoming with detecting mango trees by the said method is that the trees with touching and overlapping crowns appear as single connected object in segmentation output and hence, are detected as a single tree. This greatly affects the detection accuracy in a dense plantation/forest. This research goes a step further to address the shortcomings of the said method. We suggest a novel method to overcome this problem which can be used for instance segmentation in binary class problems. The FCN used for semantic segmentation is retrained for multi-class segmentation. In this case, $3$ classes which include mango tree crowns, boundary separating touching/overlapping crowns and background. The output of FCN trained for 3-class segmentation would contain touching/overlapping crowns separated. Subsequently, connected objects are detected and bounding boxes are drawn in original images. We prove that this method leads to an increased accuracy in detection of individual mango tree crowns and obtaining their count. We train the FCN network separately for semantic segmentation and individual tree crown detection tasks.

The rest of the paper is organized into four main sections. A description of FCNs is provided in section \ref{sec:FCNs}. The methodology used is detailed in section \ref{sec:method}. The results and inference drawn are discussed in section \ref{sec:results}. Conclusion and future scope are discussed in section \ref{sec:conclusion}.  

\section{Fully Convolutional Neural Networks (FCNs)}
\label{sec:FCNs}

In this section, the description of Fully Convolutional Neural Network (FCN), which is the basis for proposed method, is provided along with various operations involved within a FCN.

An Artificial Neural Network (ANN) is a system of interconnected neurons – inspired by the biological neural networks that constitute animal brains – that can model complicated functions. ANNs find wide range of applications in computer vision tasks such as classification, segmentation and regression.  The most basic ANN consists of three layers – input layer, hidden layer and output layer – with each layer having multiple nodes (neurons). ANNs are supervised learning algorithms that are trained with labeled data to perform specific tasks.

An ANN with one-dimensional arrangement of nodes in each layer, interconnected with nodes of adjacent layers is referred to as Multi-layer Perceptron (MLP). The Convolution Neural Network (CNN) is a class of ANN in which, instead of 1 dimensional arrangement of nodes, convolutional block is used. The convolutional block consists of convolution operation which may be followed by batch-normalization and non-linear activation. A deep CNN consists of a stack of several convolutional blocks. Deep CNNs progressively learn high-level feature representations of the input data during training. The output of convolutional block consists of two-dimensional matrices referred to as feature maps. Each feature map is obtained by performing convolution operation of a kernel/filter with the input to the convolutional block and optionally subjecting it to batch-normalization and non-linear activation operations. The feature map obtained post convolution operation is given by equation \ref{eq:eq1}.

\begin{equation}
\label{eq:eq1}
x^{l}_{i} = \sum _{j=1}^{m} x^{l-1}_{j} \circledast w_{ij}^{l} + b_{i}^{l}
\end{equation}

where $\circledast$ is a 2D-convolution operation, $x^{l}_{i}$ is the $i^{th}$ feature map output from convolution operation of $l^{th}$ convolutional block and $m$ is the number of feature maps in the input. $w_{ij}^{l}$ is one of the weight matrices producing $i^{th}$ feature map of $l^{th}$ convolution block and $b_{i}^{l}$ is the $i^{th}$ bias term of $l^{th}$ convolution block. The set of weight matrices $(j=1 \thinspace to \thinspace m)$ is referred to as kernel/filter. 

\begin{figure}[h!]
  \centering
  \includegraphics[width=0.7\textwidth, height=130pt]{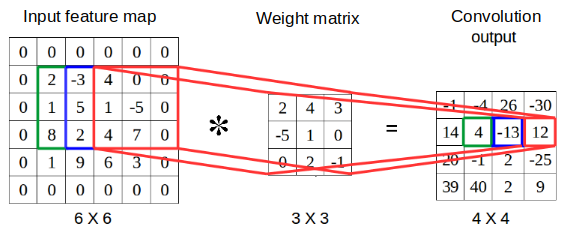}
  \caption{2D-Convolution operation}
  \label{fig:conv1}
\end{figure}

The 2D-convolution operation is the dot product of a weight matrix with the portion of feature map equal to the size of the weight matrix. The weight matrix is slid over the entire width and height of the feature map with a constant stride. The dot products at each location constitute the pixels in convolution output. This is demonstrated pictorially in figure \ref{fig:conv1}. The weight matrix $w_{ij}^{l}$ of size $3 \times 3$ is convolved with feature map $x^{l-1}_{j}$ of size $6 \times 6$. In figure \ref{fig:conv1}, the dot product of portion of feature map highlighted in red with weight matrix results in the pixel highlighted in red in the convolution output. The dot product is element wise multiplication followed by summation of results of multiplication. Each input feature map has an associated weight matrix in a kernel. The pixel wise summation of convolution outputs from input feature maps and associated weight matrices in kernel results in an output feature map. The number of feature maps output from a convolutional block $l$ is equal to number of kernels $(i=1 \thinspace to \thinspace c)$ used in that block.

The feature maps are $h \times w$ dimensional matrices where $h$ is the number of rows and $w$ is the number of columns. The output of convolutional block consisting of feature maps is of size $h \times w \times c$, where c is the number of filters. The weight matrices are generally square $n \times n$ in shape. The kernel consisting of square weight matrices is of size $n \times n \times m$, where m is the number of weight matrices also equal to number of feature maps input to convolutional block.

Convolution operation is followed by batch-normalization. Batch-normalization is known to accelerate deep neural network training by reducing internal covariate shift \citep{ioffe2015}. Besides, batch-normalization enables use of higher learning rates and regularizes the model. The mean and variance are computed about the batch, height and width dimensions of feature maps. Batch dimension refers to the number of examples used for training per iteration. Batch-normalization is achieved by element wise subtraction of mean and division of the result by standard deviation. Further, scale $\gamma$ and offset $\beta$ are used to scale and shift the normalized pixel value (equation \ref{eq:eq2}). $\gamma$ and $\beta$ are trainable parameters. For further details we refer to the original paper [17].

\begin{equation}
  \label{eq:eq2}
    \begin{split}
       \hat{p}_{ij} = (p_{ij} - \mu_B) / \sqrt{\sigma^2_B + \epsilon} \\
       p'_{ij} = \gamma \thinspace \hat{p}_{ij} + \beta 
    \end{split}
\end{equation}

where $p_{ij}$ is the pixel in a feature map, $\mu_B$ and $\sigma^2_B$ are the mini-batch mean and variance. $\hat{p}_{ij}$ is the normalized value of the pixel. $\gamma$ and $\beta$ are trainable parameters that are learned by the network. $p'_{ij}$ is the batch-normalized value of the pixel.

\begin{figure}[h!]
  \centering
  \includegraphics[width=0.65\textwidth, height=115pt]{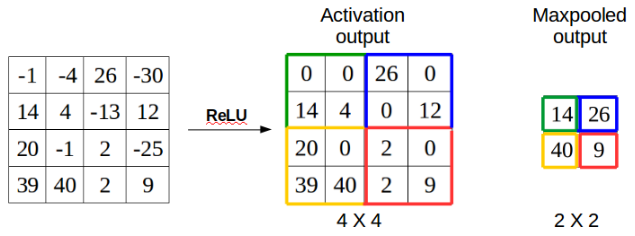}
  \caption{ReLU activation followed by max-pooling}
  \label{fig:pool}
\end{figure}

The batch-normalized feature maps are subjected to pixel wise non-linear activation. ReLU function in equation \ref{eq:eq3} is applied pixel wise. Subsequently, the feature maps are subjected to max-pooling operation. Max-pooling reduces the spatial dimensionality and complexity by extracting high-level global features from local features. It helps to train faster by reducing the number of mathematical operations as the network grows deeper while preserving essential features. The ReLU and max-pooling operations are demonstrated in figure \ref{fig:pool}. The output of ReLU activation would have all the negative values replaced by zeros. The max-pooling operation retains maximum element in a portion of feature map instead of whole portion. The size of the portion is determined by the stride. In figure \ref{fig:pool}, stride chosen is $2$. Maximum element from each $2 \times 2$ portion of the input is retained.   

\begin{equation}
\label{eq:eq3}
f(x) = max(0,x)
\end{equation}

The CNN is terminated differently for different tasks. For a classification or a regression task, the stack of convolutional blocks is followed by one or several fully connected/dense layers (MLP). The terminating fully connected layer has number of nodes equal to classification categories. For the Semantic segmentation task, the network is terminated with a convolution layer. Such a network is referred to as Fully Convolutional Neural Network (FCN). The FCN can be interpreted as transformation from image space to segmentation map space. Segmentation maps contain class of each pixel. 

\begin{figure}[h!]
  \centering
  \includegraphics[width=0.6\textwidth, height=115pt]{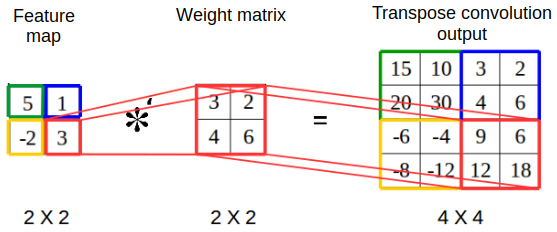}
  \caption{Transpose convolution operation}
  \label{fig:tc}
\end{figure}

Subsequent max-pooling operations reduce the size of feature maps. The feature maps have to be upscaled to obtain an output with size same as input image. Thus, FCN consists of transpose convolution operations to undo the down sampling effect of max-pooling operations. The structure of a typical FCN can be seen in figure \ref{fig:mtn}. Transpose convolution is inverse of convolution operation. In figure \ref{fig:tc}, the transpose convolution operation is demonstrated. Each pixel in the input feature map is multiplied with weight matrix and the product replaces the single pixel. The stride chosen for demonstration is $2$. If the stride chosen is less than size of weight matrix, the products of pixels and weight matrix are added in the region of overlap in the output. If the stride chosen is greater than size of weight matrix, the gaps in the output are filled with zeros. Like convolution, the kernel has number of weight matrices equal to number of input feature maps. The transpose convolution results of feature map and weight matrix pair are summed element wise to obtain an output feature map. The number of output feature maps are equal to number of kernels used.

The transpose convolution may be followed by batch-normalization, activation and convolution operations which are architecture specific. The final convolution block of FCN has number of filters equal to number of classes in case of multi-class dataset and $1$ or $2$ in case of binary-class dataset. The output of the FCN gives pixel wise class probability, indicating the chance of the pixel belonging to one of the target classes.            

\section{Methodology}
\label{sec:method}

This section describes the proposed methods for semantic segmentation and individual crown detection of mango trees. The proposed methods are based on an FCN referred to henceforth as Mango Tree Net. The various steps involved in our method: data acquisition, ground truth generation, training, inference and evaluation are discussed in this section. Further, evolution through various architectures to arrive at Mango Tree Net is also discussed. 

In the data preparation step, the ground truth images are generated for the images acquired. The ground truth image is an image that provides the reference class label of each pixel.  The images and the corresponding ground truth images are split into training and test sets. For semantic segmentation task, the Mango Tree Net is trained to classify each pixel into one of 2 classes – mango tree crown and background. Semantic segmentation is carried out using fully trained Mango Tree Net. The output of Mango Tree Net is subjected to thresholding to obtain segmentation map. For detection of individual tree-crowns, the Mango Tree Net is trained to classify each pixel into one of 3 classes – mango tree crown, boundary separating overlapping/touching crowns and background. The output of the Mango Tree Net trained for 3 class classification is a probability map. The probability map gives probability of a pixel belonging to each of the 3 classes. Segmentation map is obtained by assigning each pixel to class with highest probability. The segmentation map thus obtained would have overlapping/touching crowns separated. Connected object detection is applied on the segmentation map to detect candidate mango trees. Spurious detections due to noise in segmentation output are filtered out. Using the coordinates of connected objects detected, bounding boxes are drawn on original images. Thus, individual mango tree crowns are detected and their count is obtained.

\subsection{Data preparation}
The data preparation step involves data acquisition, ground truth generation and data pre-processing which are detailed in this subsection.

\subsubsection{Data acquisition}
An autonomous quadrotor UAV was flown over the experiment site with flight path calculated to cover the region of interest. The images of mango tree crowns were extracted from a HD video shot using a GO PRO mounted on the UAV. The images extracted lie in visual range of spectrum and have a resolution of 1080 x 1920. Videos were acquired under different lighting conditions at different times of a day during summer season. The images representing various regions of the orchard containing 6 varieties of mango trees, young and old, intermixed with other vegetation, were handpicked to prepare a diverse dataset. Images containing trees that seemed ambiguous through visual inspection were cropped to remove the ambiguity.

\subsubsection{Ground truth generation}
For ground truth creation, GIMP – an open source image manipulation tool was used. Ground truth labelling was carried out manually through visual inspection of the images. Two different sets of ground truth images were created, one for semantic segmentation and other for detection of individual crowns. In the ground truth images for semantic segmentation, pixels of mango tree crowns were marked in green and all others pixels of background class in black. In the ground truth images for individual crown detection, pixels of  mango tree crowns were marked in green, pixels of boundary separating overlapping/touching crowns in white and all other pixels of background class in black. The ground truth images (figure \ref{fig:fm}(b)) are stored as bitmap files.

The green connected regions in the ground truth for individual crown detection were bound by a box. We refer to the bounding box as annotation box. Each annotation box contains a single mango tree crown. These annotation boxes were used for the evaluation of individual crown detection and counting.
        
\subsubsection{Data pre-processing}

The ground truth images containing pixels marked with colors indicating their class are converted into forms that can be used for training. For semantic segmentation, the corresponding ground truth images are converted into pixel class maps. A pixel class map contains $1$s in place of green pixels (mango tree crowns) and $0$s in place of black pixels (background). The size of a pixel class map would be $m \times n$, where $m$ and $n$ are height and width of ground truth image. For individual crown detection, corresponding ground truth images are converted into one hot encoded labels. The one hot encoded labels would have size of $m \times n \times 3$, where $m$ and $n$ are height and width of ground truth image, and $3$ is the number of classes. One hot encoded labels are $3$ pixel class maps stacked along the last dimension. The first pixel class map contains $1$s in place of green pixels (mango tree crowns) and $0$s elsewhere. The second pixel class map contains $1$s in place of white pixels (boundary separating overlapping crowns) and $0$s elsewhere. The third pixel class map contains $1$s in place of black pixels (background) and $0$s elsewhere.

	The images and their corresponding pixel class maps / one hot labels were cropped into small patches of size $240 \times 240$ and subjected to data augmentation. The cropping was necessitated by the limitations of the available GPU (NVIDIA K$620$) and computational resources. Training a large mini-batch of images of size $1080 \times 1920$ on a GPU with $2$GB memory is not possible. For batch-normalization to be effective and training loss to decrease smoothly without fluctuations, a large mini-batch size is helpful. Cropping facilitated using a mini-batch size of $16$ for training.

	Cropping, and data augmentation through rotation and mirror about diagonal yielded a training dataset containing $8824$ image patches. The test dataset for semantic segmentation task contains $36$ RGB images of sizes $512 \times 960$ and $1080 \times 1920$. A separate test dataset with images containing a large number of overlapping/touching tree crowns was chosen for individual crown detection task. It consists of $4$ images having $297$ mango trees in total. No patches were made during testing as a complete image could be segmented in one pass. 

\subsection{Arriving at Mango Tree Net architecture}
Arriving at a suitable architecture for a given problem is a non-trivial task. A deep FCN involves a large number of hyperparameters which are fixed by experimentation. Number of convolution blocks, number of kernels within each block were fixed through Arch-1 to Mango Tree Net by experimentation. The description of observations made and the Mango Tree Net architecture are discussed in detail in this section. Table \ref{tab:archs} contains the details of each of the architectures considered and Table \ref{tab:params} contains the number of trainable parameters in each architecture. 

\begin{table}[h!]
  \centering
  \begin{tabular}{|c|c|c|c|}\hline
    \textbf{Arch-1}     & \textbf{Arch-2}     & \textbf{Arch-3} & \textbf{Mango Tree Net}\\ \hline
    Conv 3x3 / 8 & Conv 3x3 / 8 & Conv 3x3 / 16 & Conv 3x3 / 16\\ \hline
    Conv 5x5 / 8 & Conv 5x5 / 8 & Conv 5x5 / 16 & Conv 5x5 / 16\\ \hline
    Max-Pooling & Max-Pooling & Max-Pooling & Max-Pooling\\ \hline
    Conv 7x7 / 16 & Conv 7x7 / 16 & Conv 7x7 / 32 & Conv 7x7 / 32\\ \hline
    Max-Pooling & Max-Pooling & Max-Pooling & Max-Pooling\\ \hline
    Conv 5x5 / 32 & Conv 5x5 / 32 & Conv 5x5 / 64 & Conv 5x5 / 64\\ \hline
    Transpose Conv 2x2 / 32 & Transpose Conv 2x2 / 32 & Transpose Conv 2x2 / 64 & Max-Pooling\\ \hline
	Transpose Conv 2x2 / 16 & Conv 7x7 / 16 & Conv 7x7 / 32 & Conv 5x5 / 128\\ \hline
	Conv 1x1 / 1 & Transpose Conv 2x2 / 16 & Transpose Conv 2x2 / 32 & Transpose Conv 2x2 / 128\\ \hline
	& Conv 5x5 / 8 & Conv 5x5 / 16 & Conv 5x5 / 64\\ \hline
	& Conv 3x3 / 8 & Conv 3x3 / 16 & Transpose Conv 2x2 / 64 \\ \hline
	& Conv 1x1 / 1 & Conv 1x1 / 1 & Conv 7x7 / 32\\ \hline
	& & & Transpose Conv 2x2 / 32 \\ \hline
	& & & Conv 5x5 / 16\\ \hline
	& & & Conv 3x3 / 16\\ \hline
	& & & Conv 1x1 / 1 / 3\\ \hline
	\end{tabular}
  \newline
  \caption{Various architectures considered}
  \label{tab:archs}
\end{table}

\begin{table}[h!]
  \centering
    \begin{tabular}{|c|c|c|c|c|}\hline
	  \textbf{Architecture} 				& Arch-1 & Arch-2 & Arch-3 & Mango Tree Net\\ \hline
	  \textbf{No. of Trainable Parameters} & 0.027 & 0.054 & 0.219 & 0.663\\ \hline
	\end{tabular}

  \caption{Number of trainable parameters(in millions) in architectures considered}
\label{tab:params}
\end{table}

\begin{figure}[h!]
\centering
  \begin{minipage}{0.49\linewidth}
  \def\svgwidth{\columnwidth}
    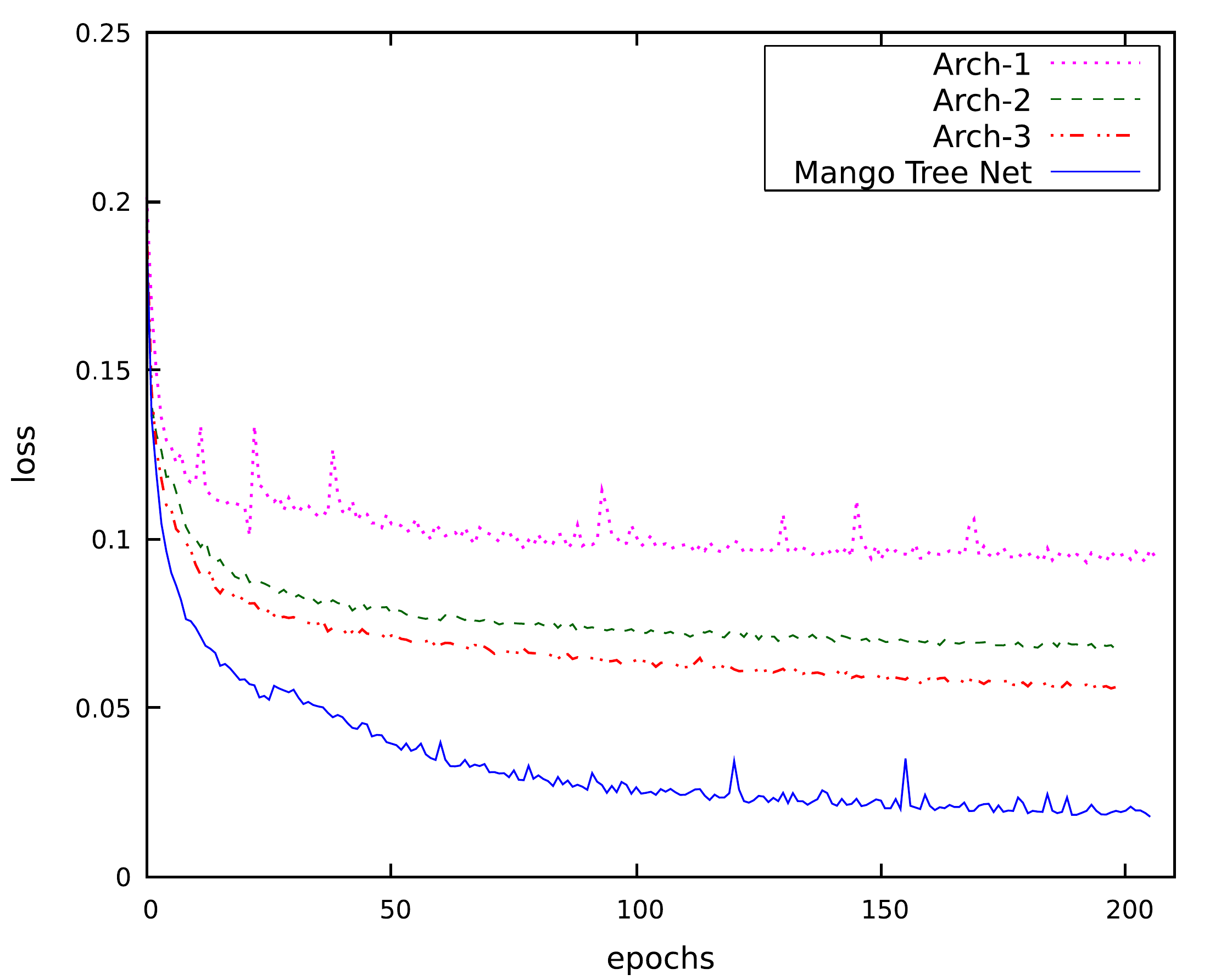
  \caption{Training loss plots of architectures considered \newline }
  \label{fig:loss}
  \end{minipage}
  \begin{minipage}{0.49\linewidth}
  \def\svgwidth{\columnwidth}
    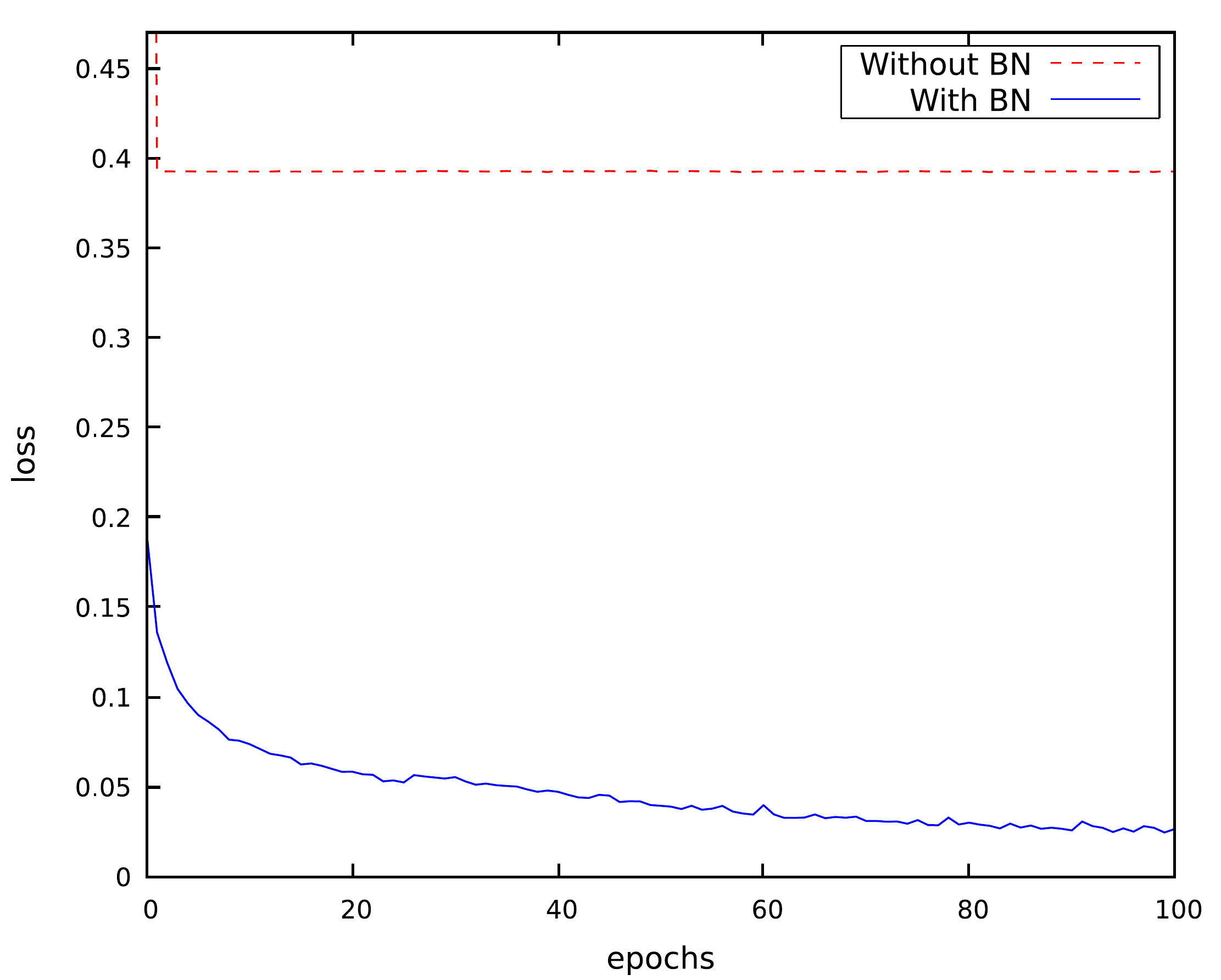
  \caption{Comparison of loss plots of Mango Tree Net with and without batch-normalization}
  \label{fig:loss_ubn}
  \end{minipage}
\end{figure}

The operation mentioned as Conv in table \ref{tab:archs} refers to 2D-convolution followed by batch-normalization and ReLU activation everywhere except the final block. After the final convolution operation in each architecture sigmoid activation function is used. Each of the architectures was trained on complete training dataset. It was observed that arch-1 was largely underfitting on the data. This can be seen from the loss curve in figure \ref{fig:loss}. The average loss value for each epoch is plotted. The loss curve of arch-1 decreases for a initial few epochs before hitting a plateau. The predictions made on fully trained arch-1 had large number of false positives, indicating the network could not very well differentiate between mango trees and other vegetation. The same can be observed from the heat maps in figure \ref{fig:hm} in the results section. Consequently, to increase the learning capacity convolutional blocks were added after transpose convolution operations. The improvement caused by this addition is seen in the loss curve of arch-2. The loss decreases sharply for initial epochs before the decrease is tapered off. The loss curve almost flattens around 0.08 with increase in the number of epochs. This indicates an under fit on the data and a need to further increase the learning capacity. Keeping the number of convolution and transpose convolution blocks intact, the number of kernels were doubled in arch-3 to increase the learning capacity. The loss curve of arch-3 follows the trend of loss curve of arch-2, indicating a underfit. The heat maps for arch-2 and arch-3 show high probability for some pixels containing non-mango vegetation. Further convolution, max-pooling, transpose convolution blocks were added as in Table A to increase the learning capacity. We refer to this architecture that we have finally converged on as Mango Tree Net. The loss curve obtained using this architecture is the one that is more desirable. The curve does not hit a plateau abruptly, but decreases gradually with increasing epochs indicating no under fit and successful convergence of the network. The importance of batch-normalization in Mango Tree Net can be seen in figure \ref{fig:loss_ubn}. The network does not converge without batch-normalization. It can be seen that the loss curve remains stagnant pegged around $0.39$. 

The comparison of the results obtained on testing set using each architecture are discussed in the section \ref{sec:results}. It can be seen from the heat maps in figure \ref{fig:hm} that there is a gradual improvement of performance across architectures. This improvement can be attributed to the increasing number of trainable parameters from one architecture to another as seen in Table \ref{tab:params}. The increase in the number of trainable parameters results in increased learning capability and thus, better performance.
	
\subsubsection{Anatomy of Mango Tree Net}

\begin{figure}[h!]
  \centering
  \def\svgwidth{\columnwidth}
    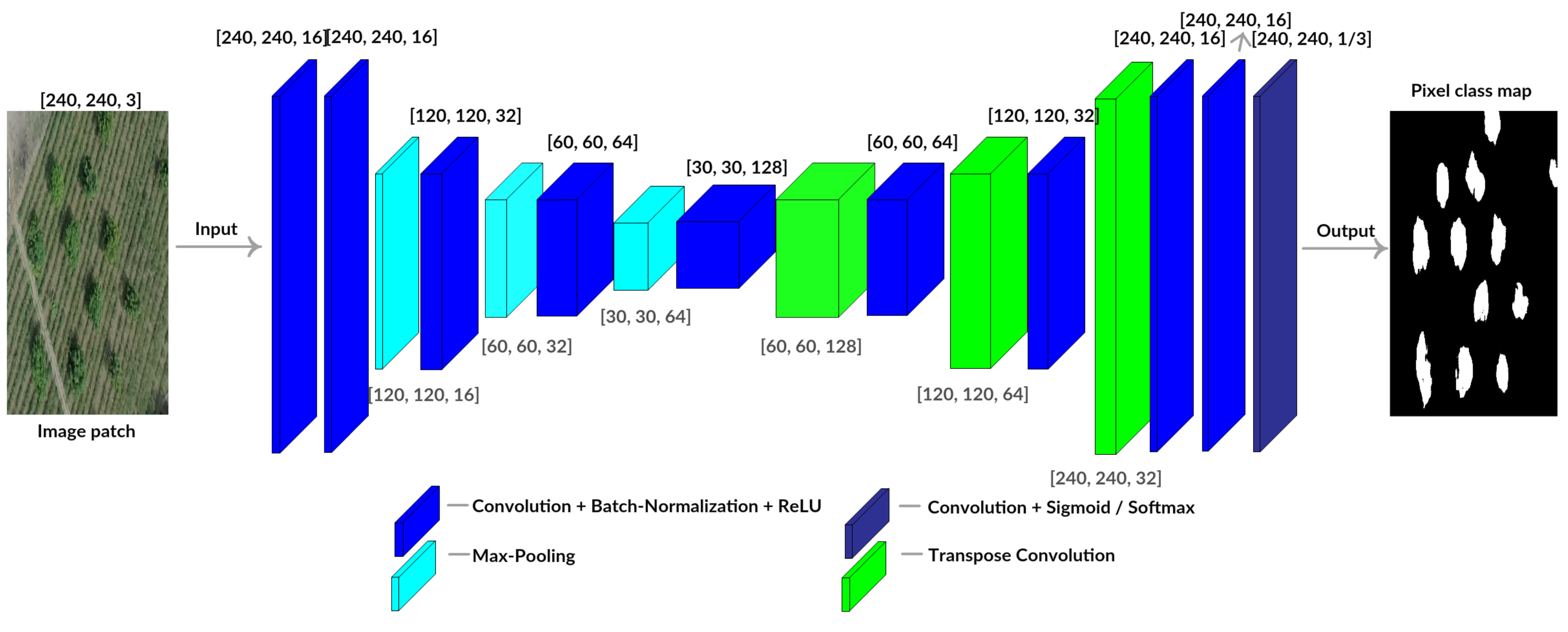
  \caption{Architecture of Mango Tree Net}
  \label{fig:mtn}
\end{figure}

The block diagram of Mango Tree Net is depicted in figure \ref{fig:mtn}. Different blocks in the Mango Tree Net are depicted in different colors. Convolutional blocks are depicted in blue, max-pooling operations are depicted in cyan, transpose convolution operations are depicted in green and the terminating convolutional block is depicted in purple. The network has $9$ convolution blocks consisting of convolution operation followed by batch-normalization and non-linear activation using ReLU and the terminating convolution block consisting of convolution operation followed by sigmoid or softmax activation. There are $3$ max-pooling operations and $3$ transpose convolution operations to undo the effect of down sampling of max-pooling operations. The symmetric network like Mango Tree Net is referred to as U-shaped network or hour glass shaped network. \citet{olaf2015} used u-shaped network with skip connections for bio-medical segmentation. Kernels of sizes $3 \times 3$, $5 \times 5$, $7 \times 7$ are used in convolution operation. The rationale behind using kernels of bigger size is to have bigger receptive field. Bigger receptive field helps to exploit global pixel dependencies. Each convolution kernel has an associated bias term. The number of kernels and bias terms in the first five convolution operations are $16$, $16$, $32$, $64$ and $128$. The network being symmetric they are gradually reduced. The number of kernels in last five convolution operations and bias terms are $64$, $32$, $16$, $16$, $1or3$. For semantic segmentation task, the final convolution block has $1$ kernel and is terminated with sigmoid activation. For individual crown detection task, the final convolution block has $3$ kernels and is terminated with softmax activation. The size of outputs of convolutional blocks are mentioned in black text on top of respective block in figure \ref{fig:mtn}. The size of outputs of max-pool and transpose convolution operations are mentioned in gray text at the bottom of respective operations. The sizes are mentioned as $[h, w, c]$, where $h$ is height, $w$ is width, $c$ is number of feature maps/filters.

\subsection{Training}

The input to the FCN based model is an image, a $m \times n \times 3$ dimensional rank-$3$ tensor. The output in case of semantic segmentation task is a $m \times n$ dimensional rank-$2$ tensor. The output in case of individual crown detection is a $m \times n \times 3$ dimensional rank-$3$ tensor. Let $w_{j}^{l}$ be the $j^{th}$ kernel in $l^{th}$ convolutional block, $b_{j}^{l}$ be the $j^{th}$ bias term in $l^{th}$ convolutional block, $X$ be the input image and $Y$ be the output class probability map. The FCN based model Mango Tree Net $(F)$ can be interpreted as a transformation $F_{w_{j}^{l}, b_{j}^{l}}$: $X \longrightarrow Y$ with $w_{j}^{l}$, $b_{j}^{l}$ as parameters. The objective is to choose $w_{j}^{l}$ and $b_{j}^{l}$ such that the loss is minimum. The loss function for semantic segmentation task is given by equation \ref{eq:sg_loss}. The sigmoid activation is applied to output of final convolution layer (equation \ref{eq:sigmoid}) and loss is computed pixel wise for Mango Tree Net output using equation \ref{eq:sg_loss}. The pixel wise loss is then averaged over all pixels and mini-batch of outputs. The averaged loss computed is used to update the weights and biases using backpropagation algorithm and a suitable optimizer. The process is repeated until the loss converges.

\begin{equation}
\label{eq:sigmoid}
\hat{y} = \frac{1}{1+\exp{(-y_{p})}}
\end{equation}

\begin{equation}
\label{eq:sg_loss}
l_{sg} = - y\log{\hat{y}} - (1-y)\log{(1-\hat{y})}
\end{equation}

where $y_{p}$ is pixel value in output of final convolution, $\hat{y}$ is sigmoid activated pixel value, $y$ is the pixel value in ground truth pixel class map and $l_{sg}$ is the sigmoid cross entropy loss.

The network is trained separately for the individual crown detection task. The output of final convolution layer is subjected to softmax activation. The $m \times n \times 3$ dimensional output of the final convolution contains class scores along the last dimension. Softmax is applied along last dimension to normalize the class scores using equation \ref{eq:softmax}. The normalized class scores are used to evaluate pixel wise cross entropy loss given by equation \ref{eq:sm_loss}. Due to large class imbalance between boundary class pixels and other class pixels, weighted cross entropy loss is used. A factor of $60$ is multiplied to the boundary pixel loss terms. The large factor multiplied with loss is punitive and forces the model to do better. In the absence of weighted loss, the loss due to incorrect classification of boundary pixels would be insignificant and there would be no scope for model to better. The pixel wise loss is then averaged over all pixels and mini-batch of outputs. The backpropagation algorithm and a suitable optimizer are used to update weights and biases until the loss converges.

\begin{equation}
\label{eq:softmax}
\hat{y}_{s_{i}} = \frac{\exp{(s_{i})}}{\sum_{j}{\exp{(s_{j})}}}
\end{equation}

\begin{equation}
\label{eq:sm_loss}
l_{sm} = - \sum_{i} \textit{w}_{i} y_{i} \log{(\hat{y}_{s_{i}})}
\end{equation}

where $s_{i}$ is the pixel class score of $i^{th}$ class in output of final convolution, $\hat{y}_{s_{i}}$ is the normalized or softmax activated pixel class score, $y_{i}$ is value of $i^{th}$ class in one hot encoded ground truth and $\textit{w}_{i}$ is the weight factor associated with $i^{th}$ class. The weight factors are $60,\thinspace 1,\thinspace 1$ for boundary, mango tree and background classes. $l_{sm}$ is cross entropy loss.

The network is trained on a NVIDIA K$620$ GPU, for $200$ epochs, for both the tasks separately. One round of training with entire data in training set is called an epoch. Each epoch has $N/B$ iterations, where $N$ is number of image patches in the training set and $B$ is the number of image patches in a mini-batch. Every iteration, a mini-batch of image patches are passed through the network, the loss is computed for the output, gradients are computed using backpropagation algorithm and weights and biases are updated using gradients by adam optimizer. We leave the intricate details of backpropagation and adam optimizer to the extensive literature available.

\begin{figure}[h!]
\centering
\begin{tabular}{ccc}
  \includegraphics[width=0.30\textwidth]{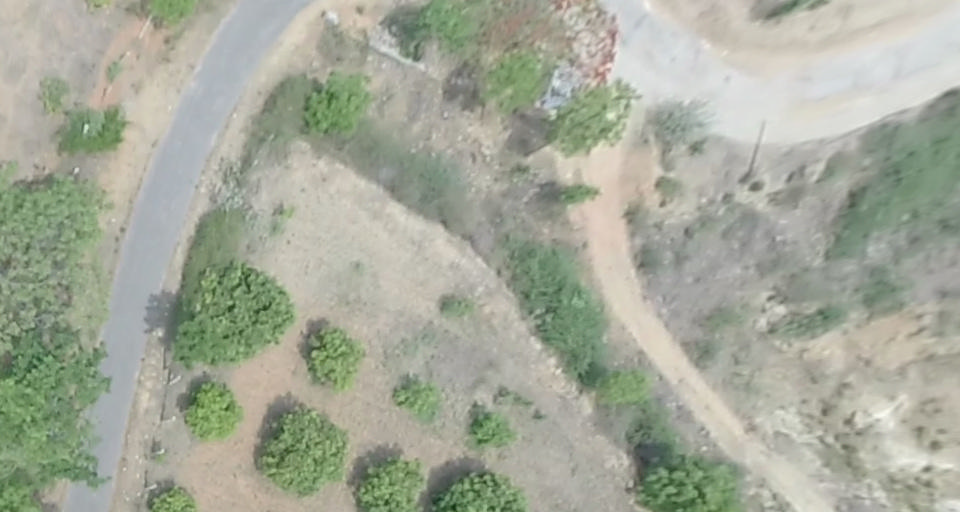} &   \includegraphics[width=0.30\textwidth]{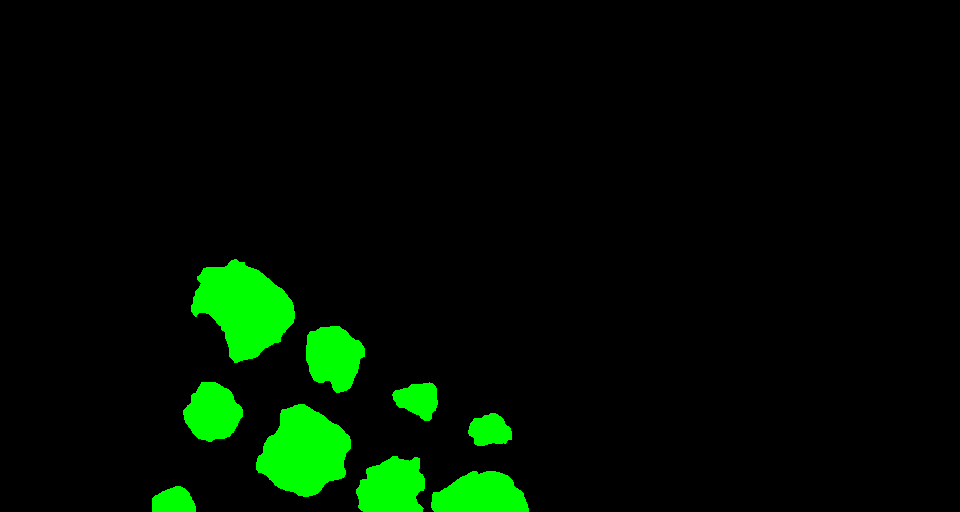} &   \includegraphics[width=0.30\textwidth]{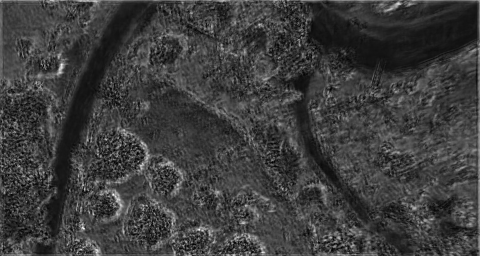} \\
(a) & (b) & (c)\\[6pt]
  \includegraphics[width=0.30\textwidth]{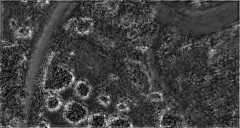} &   \includegraphics[width=0.30\textwidth]{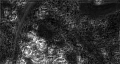} &   \includegraphics[width=0.30\textwidth]{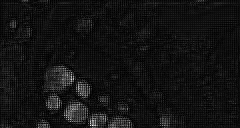} \\
(d) & (e) & (f)\\[6pt]
  \includegraphics[width=0.30\textwidth]{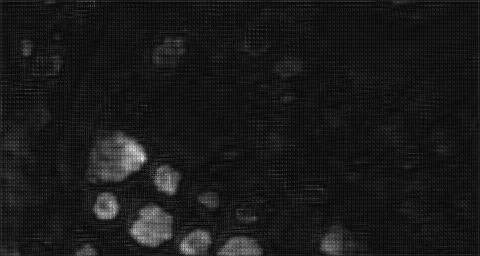} &   \includegraphics[width=0.30\textwidth]{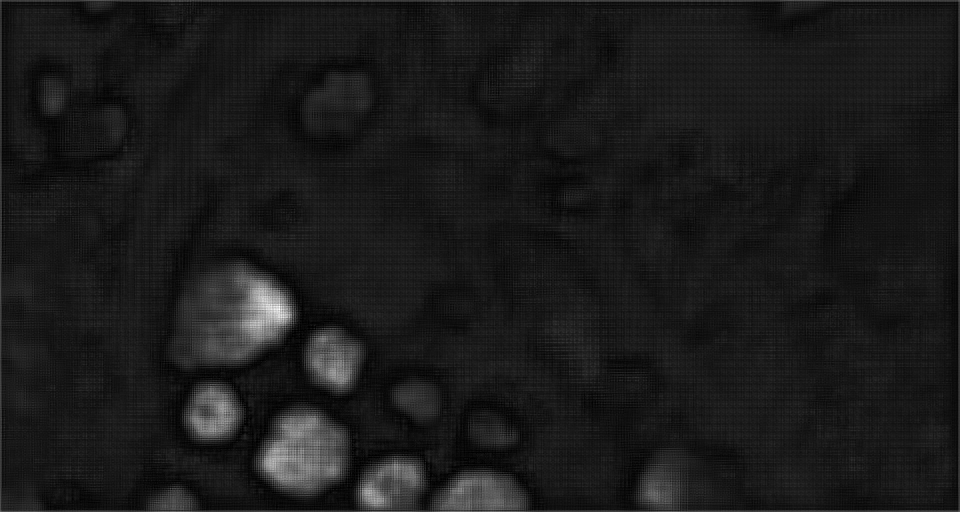} &   \includegraphics[width=0.30\textwidth]{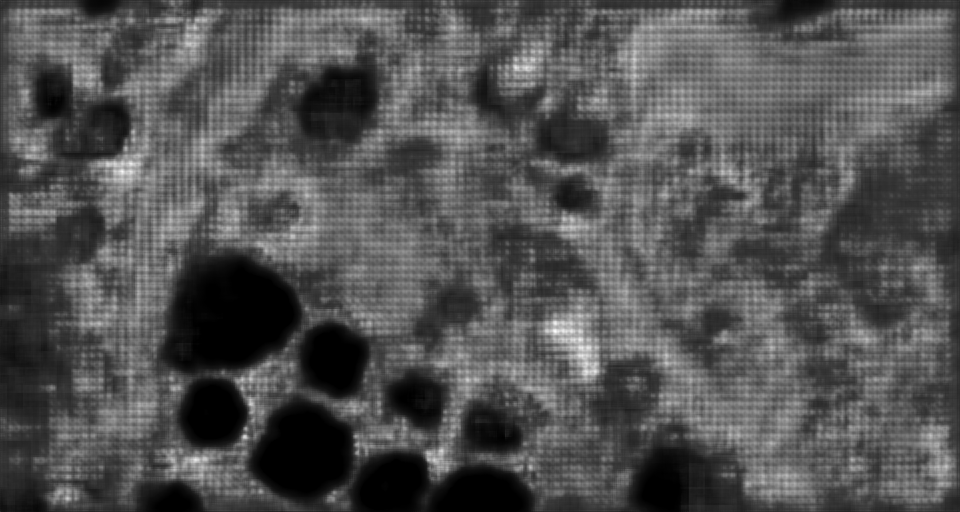} \\
(g) & (h) & (i)\\[6pt]
\end{tabular}
\caption{Test image, corresponding ground truth and averaged feature maps from intermediate convolutional blocks}
\label{fig:fm}
\end{figure}

\subsection{Visualization of intermediate feature maps}

The images in figure \ref{fig:fm} (c) - (i) are of averaged feature maps from the intermediate convolutional blocks $3$ to $9$ in Mango Tree Net. Figure \ref{fig:fm} (a) and (b) are test image and the corresponding ground truth image. Each convolution block outputs feature maps equal to number of convolution filters used. The averaged feature map is obtained by averaging the values at a given pixel position in all feature maps. The feature maps help us visualize the features that become more pronounced in output of each convolutional block. In \ref{fig:fm} (c), the periphery of mango trees are highlighted which become prominent in \ref{fig:fm} (d). In \ref{fig:fm} (e), the mango tree crowns are highlighted. In \ref{fig:fm} (f), (g) and (h) the back ground is suppressed and the tree crowns are highlighted with well defined boundaries. In \ref{fig:fm} (i), the background is prominent and mango tree crowns are suppressed. It can be seen that the resolution decreases and then increases. This is due to max-pooling and transpose convolution operations. The roads and other vegetation visible in initial average feature maps disappear gradually. Some vegetation with features similar to mango trees are visible in the initial feature maps. This is because the network learns to identify low-level features in the initial convolutional blocks and more high-level features as the network grows deeper. Hence, vegetation that appear visually similar to mango tree crowns appear in initial feature maps but disappear later. The segmentation outputs overlaid on the respective original images can be seen in figure \ref{fig:so}.

\subsection{Testing}
\label{sec:testing}

During testing phase for semantic segmentation task, each test image is passed through the Mango Tree Net. Sigmoid activation is applied pixel wise to final convolution output. Pixel wise threshold given by equation \ref{eq:thres} is applied to the  activation output. The result obtained is a segmentation map with $0$’s representing background pixels and $1$’s representing mango tree pixels. The semantic segmentation results are evaluated for performance using methods described ahead.

\begin{equation}
\label{eq:thres}
s_{ij}=\begin{cases} 
1, & \text{if } p_{ij}>0.6\\
0, & \text{if } p_{ij} \le 0.6
\end{cases}
\end{equation}

where $p_{ij}$ is the real value of pixel in range $0$-$1$ prior to thresholding and $s_{ij}$ is binary value of pixel post thresholding. The threshold chosen is 0.6 meaning pixels with probability greater than 0.6 are assigned mango tree class. 

During testing phase for individual crown detection task, each test image is passed through Mango Tree Net trained for 3-class segmentation. Softmax activation is applied to each pixel along the last dimension of final convolution output. Each pixel is assigned the class with the highest probability in softmax output. The result obtained is a segmentation map with $0$’s representing background pixels, $1$’s representing mango tree pixels and $2$’s representing boundary pixels. In the segmentation map obtained the pixels classified as boundary are set as background. The resulting map contains only $0$’s and $1$’s. The segmentation map when interpreted as a binary image contains mango tree regions as white blobs / connected objects. Connected objects detection is applied to the segmentation map to obtain coordinates of mango tree regions. Using the coordinates bounding boxes are drawn around mango trees in the original image. This is because detecting a connected object in semantic segmentation output is equivalent to detecting a mango tree in the original image. 

The individual mango tree crown detections thus obtained might contain spurious detections. This is due to the presence of noise in segmentation output. To remove spurious detections, connected objects with size less than certain threshold are omitted. The threshold for size of connected object is chosen such that it is smaller than the smallest tree in the training data. In this case, all connected objects with size less than 600 connected pixels are omitted. 

\subsection{Performance evaluation}

The results of semantic segmentation and individual tree crown detection are evaluated using standard metrics of evaluation namely precision, recall, f1-score and accuracy. The equations \ref{eq:em} $(a)-(d)$ provide the definition of each of the metrics. 

\begin{subequations}
\label{eq:em}
\begin{align}
Precision &= \frac{TP}{TP + FP} \\
Recall &= \frac{TP}{TP + FN} \\
F1-score &= \frac{2 * P * R}{P + R}\\
Accuracy &= \frac{N_{c}}{N}
\end{align}
\end{subequations}

where $TP$ is the number of true positives, $TN$ is the number of true negatives, $FP$ is the number of false positives, $N$ is the total number of entities and $N_{c}$ is the total number of entities correctly predicted. 

True positive is an entity belonging to the positive class classified as positive class. False positive is an entity belonging to the negative class classified as positive class. False negative is an entity belonging to the positive class classified as negative class. The entity in case of semantic segmentation is a pixel and performance evaluation is done at pixel level. The entity in case of individual crown detection is a tree and performance evaluation is done at tree level. Precision indicates fraction of true positive entities correctly classified among entities classified as positive. Recall indicates fraction of true positive entities among actual number of positive entities. F1-score is the harmonic mean of precision and recall that indicates the balance between precision and recall metrics. Accuracy indicates fraction of entities - both positive and negative - correctly classified. In this work, mango pixels/trees are positive class, and, background pixels and non-mango trees are negative class.

\section{Results}
\label{sec:results}

In this section, the performance of the proposed methods for semantic segmentation and individual tree crown detection tasks are discussed. In-depth analysis and visual interpretation of the results are provided. The images in test dataset were acquired using a RGB camera mounted on UAV. Images taken at varying scale, terrain, lighting conditions and surrounding vegetation are used to demonstrate the robustness of Mango Tree Net. The results section is organized as follows: In section \ref{sbsec:ssr}, performances of considered architectures on semantic segmentation task are compared and robustness of proposed architecture is demonstrated.  In section \ref{sbsec:icd}, the performance of the proposed method for individual tree crown detection task is discussed.

\subsection{Semantic segmentation results and visualization}
\label{sbsec:ssr}
\begin{figure}
\centering
  \begin{minipage}[h]{1.0\linewidth}
     \centering
     \includegraphics[width=0.15\textwidth, height=0.15\textwidth]{images/30.jpg} \hspace{0.8mm}\includegraphics[width=0.15\textwidth, height=0.15\textwidth]{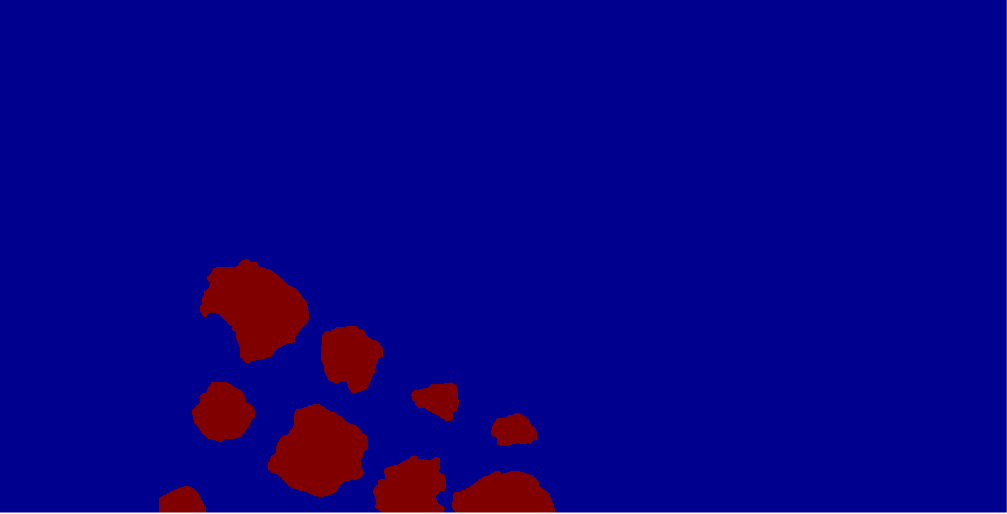} \hspace{0.8mm}\includegraphics[width=0.15\textwidth, height=0.15\textwidth]{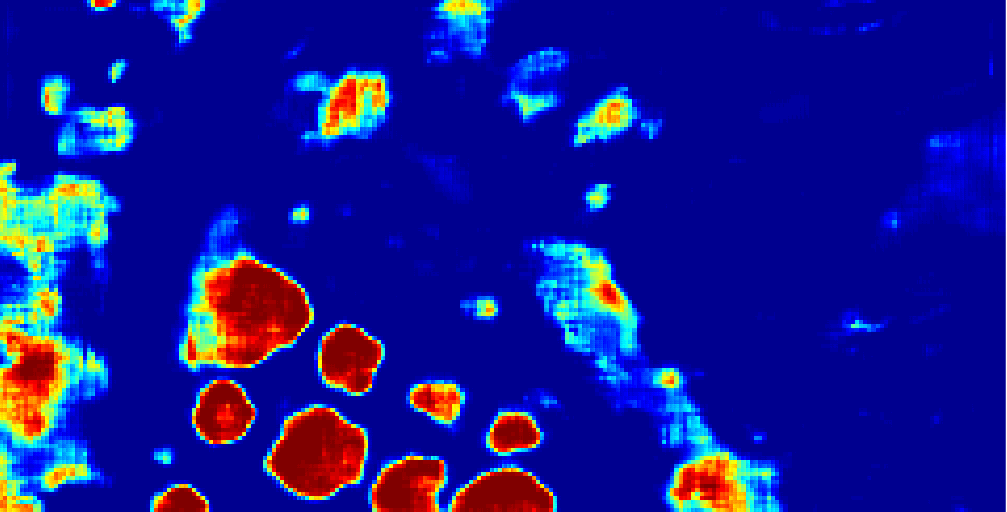} \hspace{0.8mm}\includegraphics[width=0.15\textwidth, height=0.15\textwidth]{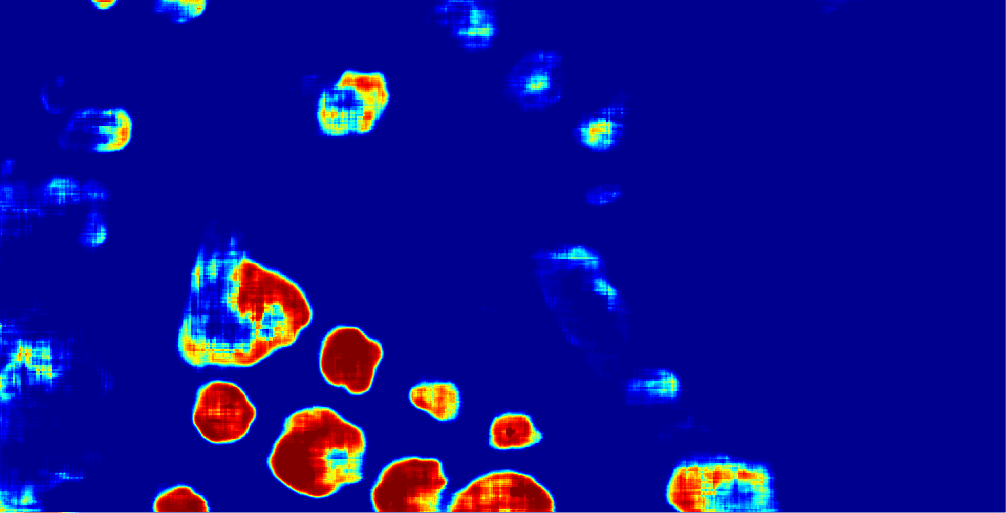} \hspace{0.8mm}\includegraphics[width=0.15\textwidth, height=0.15\textwidth]{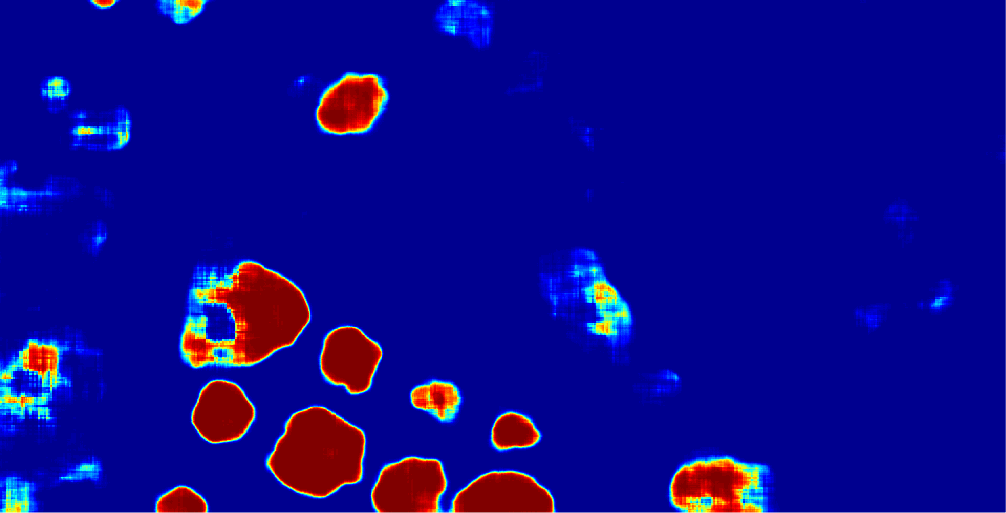} \hspace{0.8mm}\includegraphics[width=0.15\textwidth, height=0.15\textwidth]{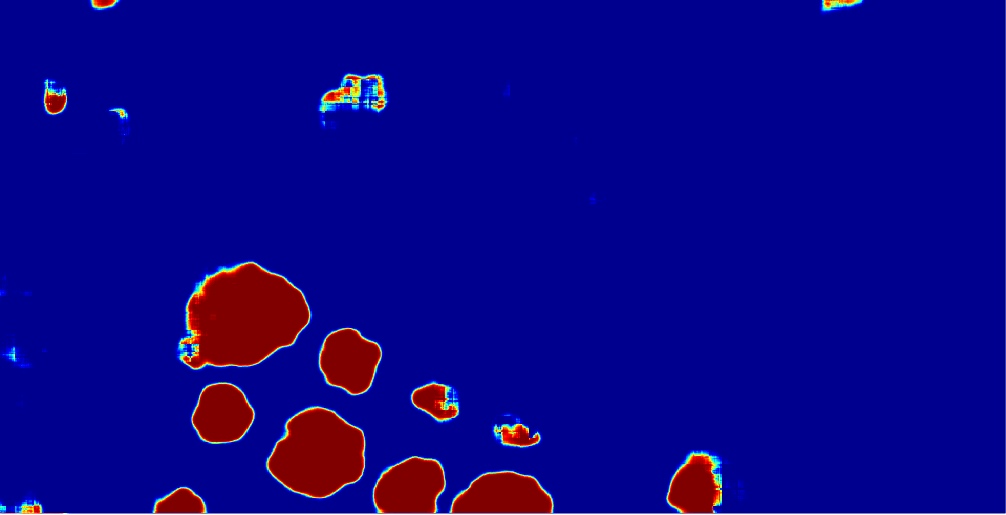} \vspace{1mm}\end{minipage}
    \begin{minipage}[h]{1.0\linewidth}
     \centering
     \includegraphics[width=0.15\textwidth, height=0.15\textwidth]{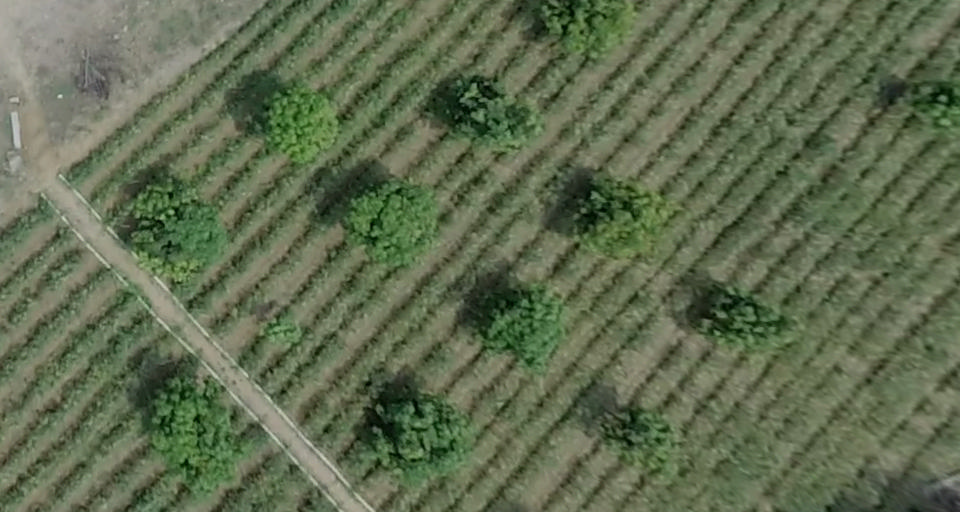} \hspace{1mm}\includegraphics[width=0.15\textwidth, height=0.15\textwidth]{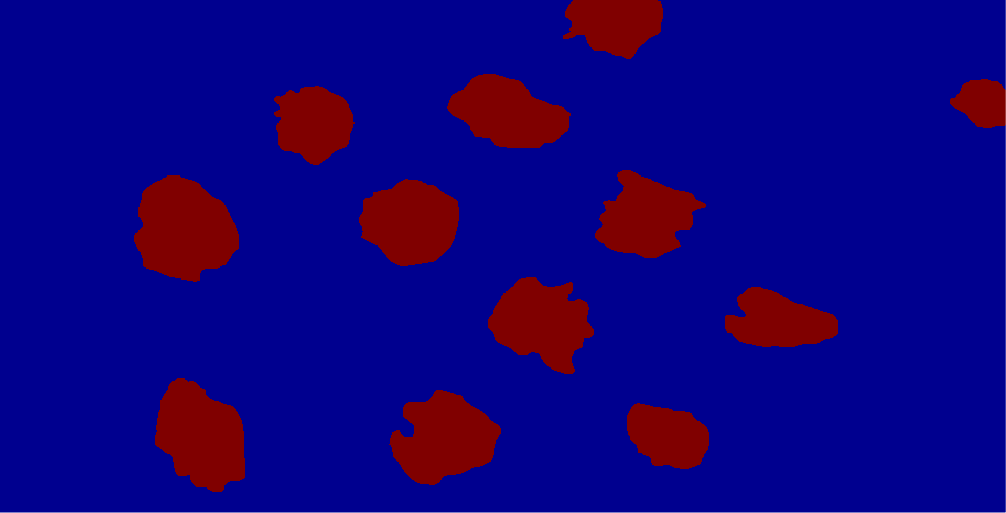} \hspace{1mm}\includegraphics[width=0.15\textwidth, height=0.15\textwidth]{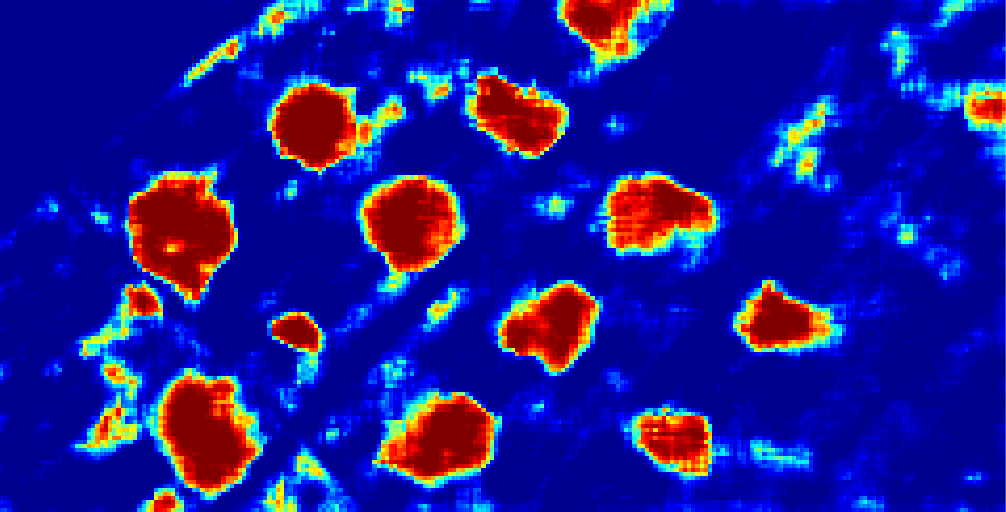} \hspace{1mm}\includegraphics[width=0.15\textwidth, height=0.15\textwidth]{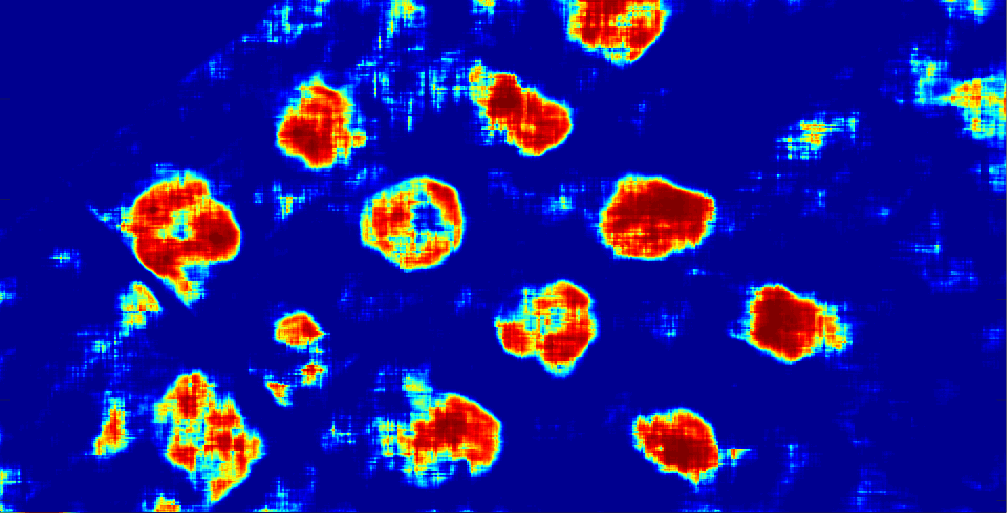} \hspace{1mm}\includegraphics[width=0.15\textwidth, height=0.15\textwidth]{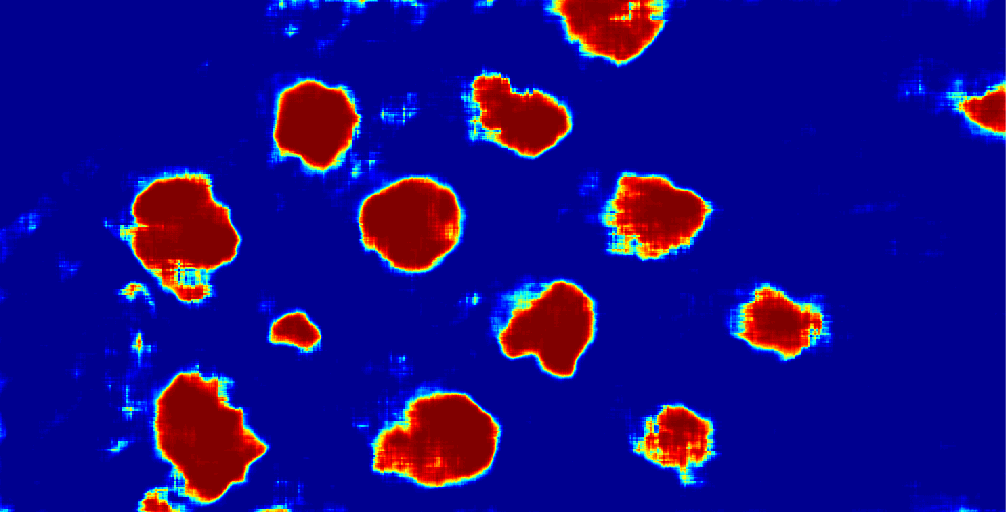} \hspace{1mm}\includegraphics[width=0.15\textwidth, height=0.15\textwidth]{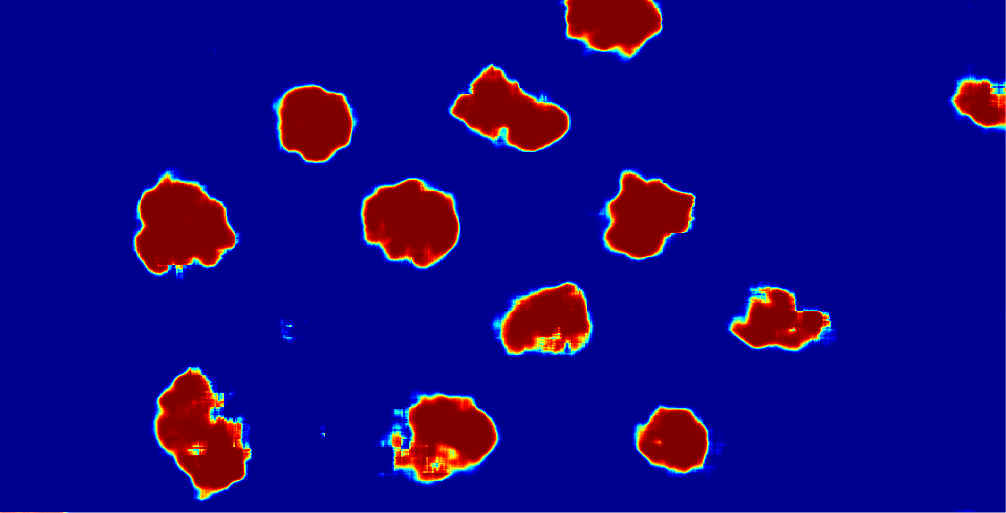} \vspace{1mm}\end{minipage}
    \begin{minipage}[h]{1.0\linewidth}
     \centering
     \includegraphics[width=0.15\textwidth, height=0.15\textwidth]{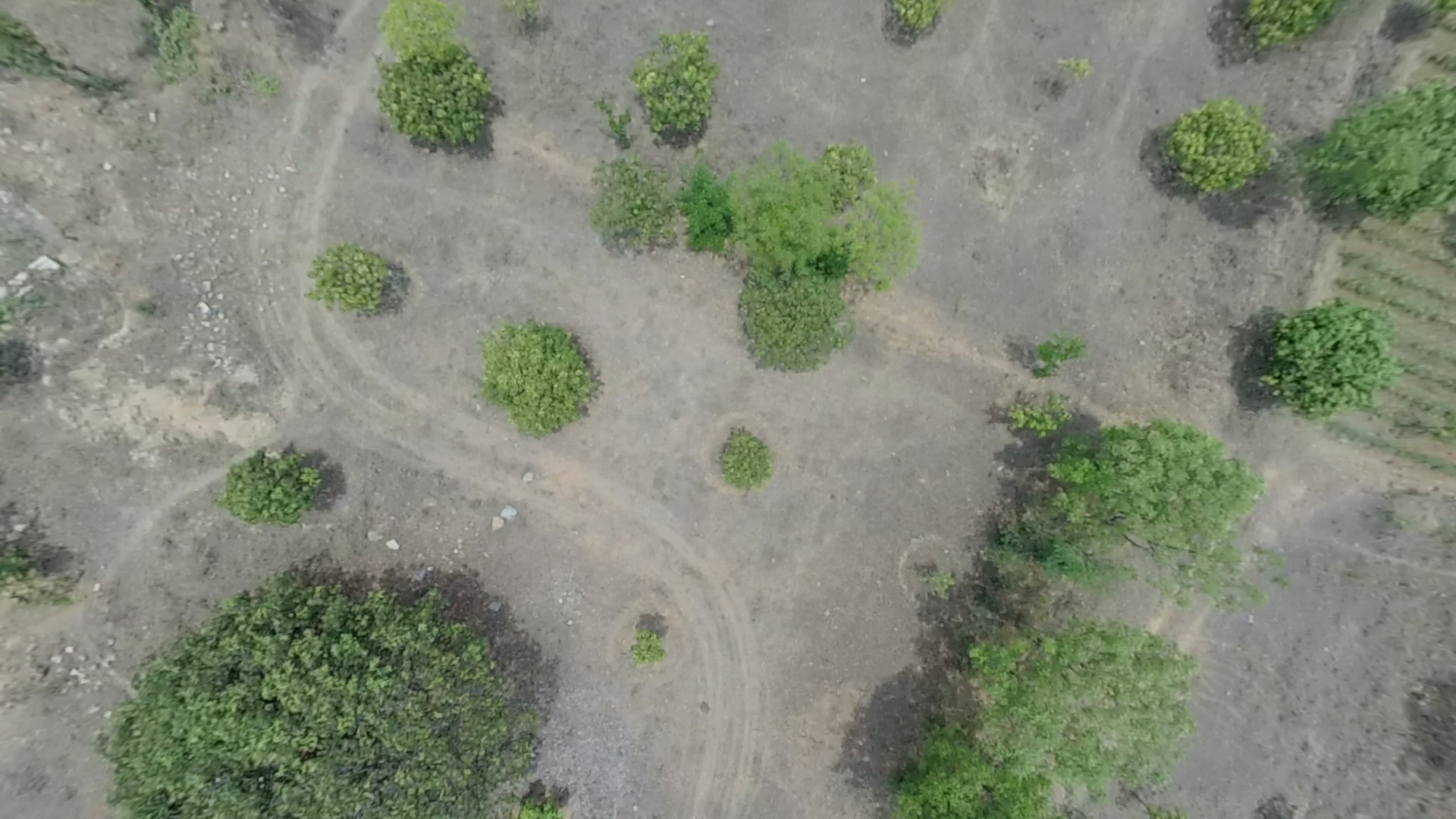} \hspace{1mm}\includegraphics[width=0.15\textwidth, height=0.15\textwidth]{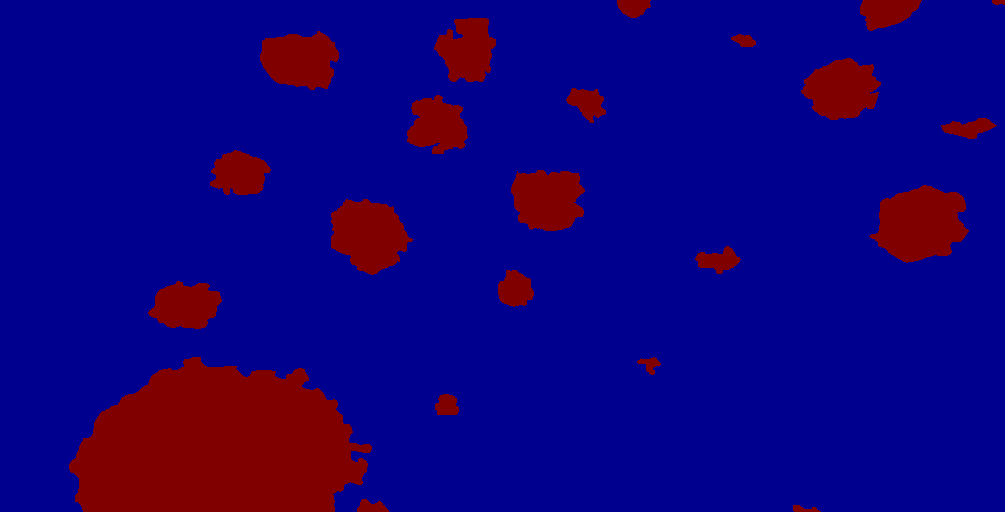} \hspace{1mm}\includegraphics[width=0.15\textwidth, height=0.15\textwidth]{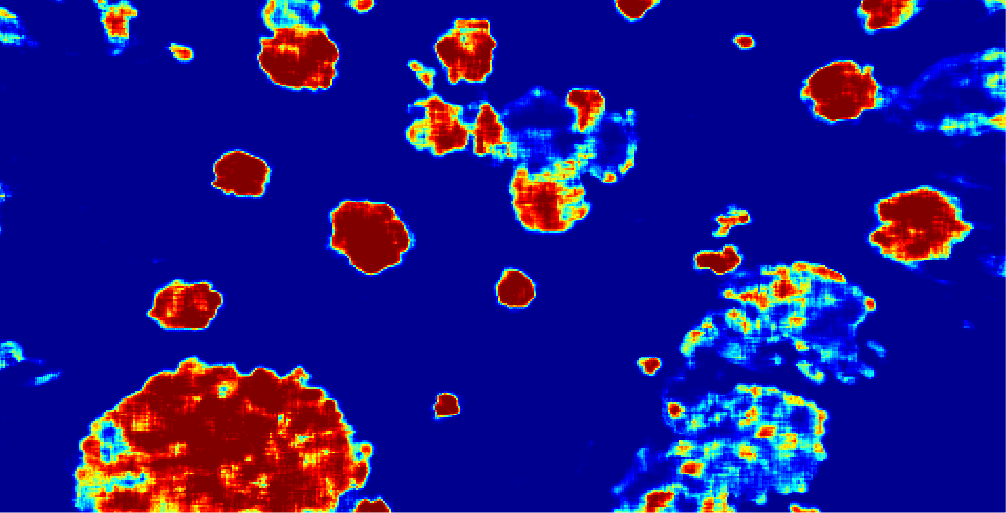} \hspace{1mm}\includegraphics[width=0.15\textwidth, height=0.15\textwidth]{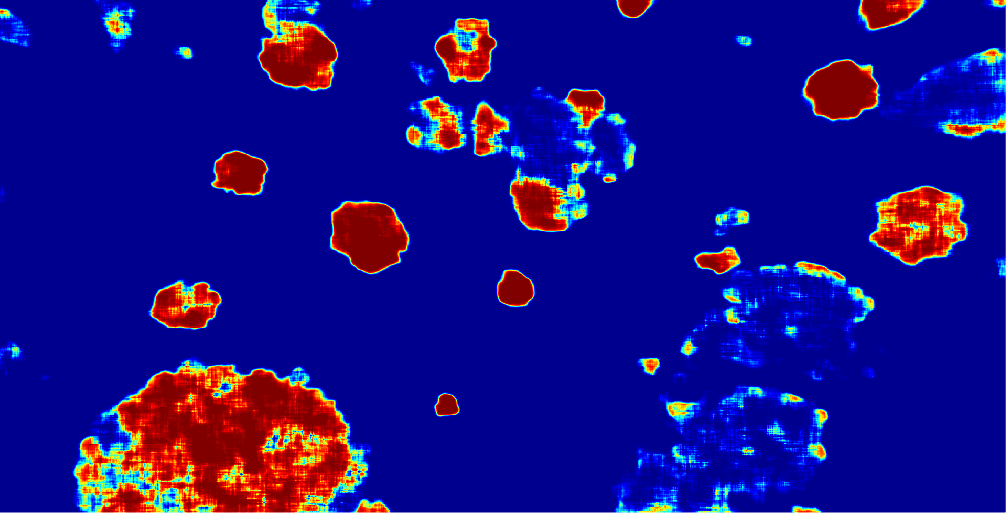} \hspace{1mm}\includegraphics[width=0.15\textwidth, height=0.15\textwidth]{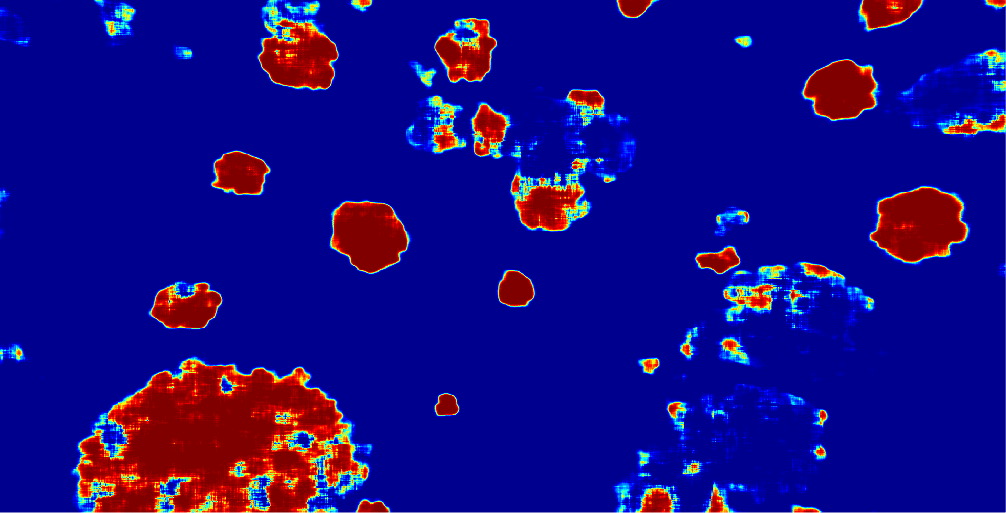} \hspace{1mm}\includegraphics[width=0.15\textwidth, height=0.15\textwidth]{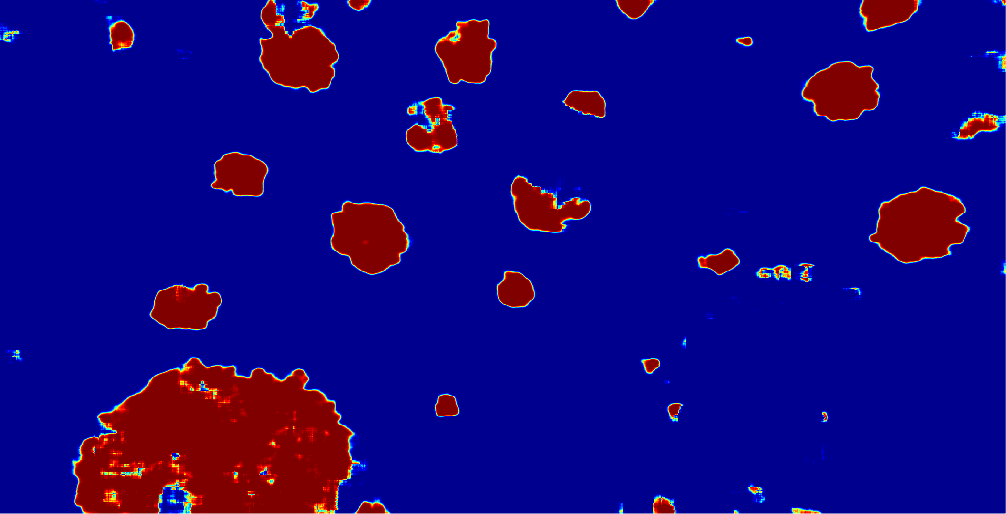} \vspace{1mm}\end{minipage}
    \begin{minipage}[h]{1.0\linewidth}
     \centering
     \includegraphics[width=0.15\textwidth, height=0.15\textwidth]{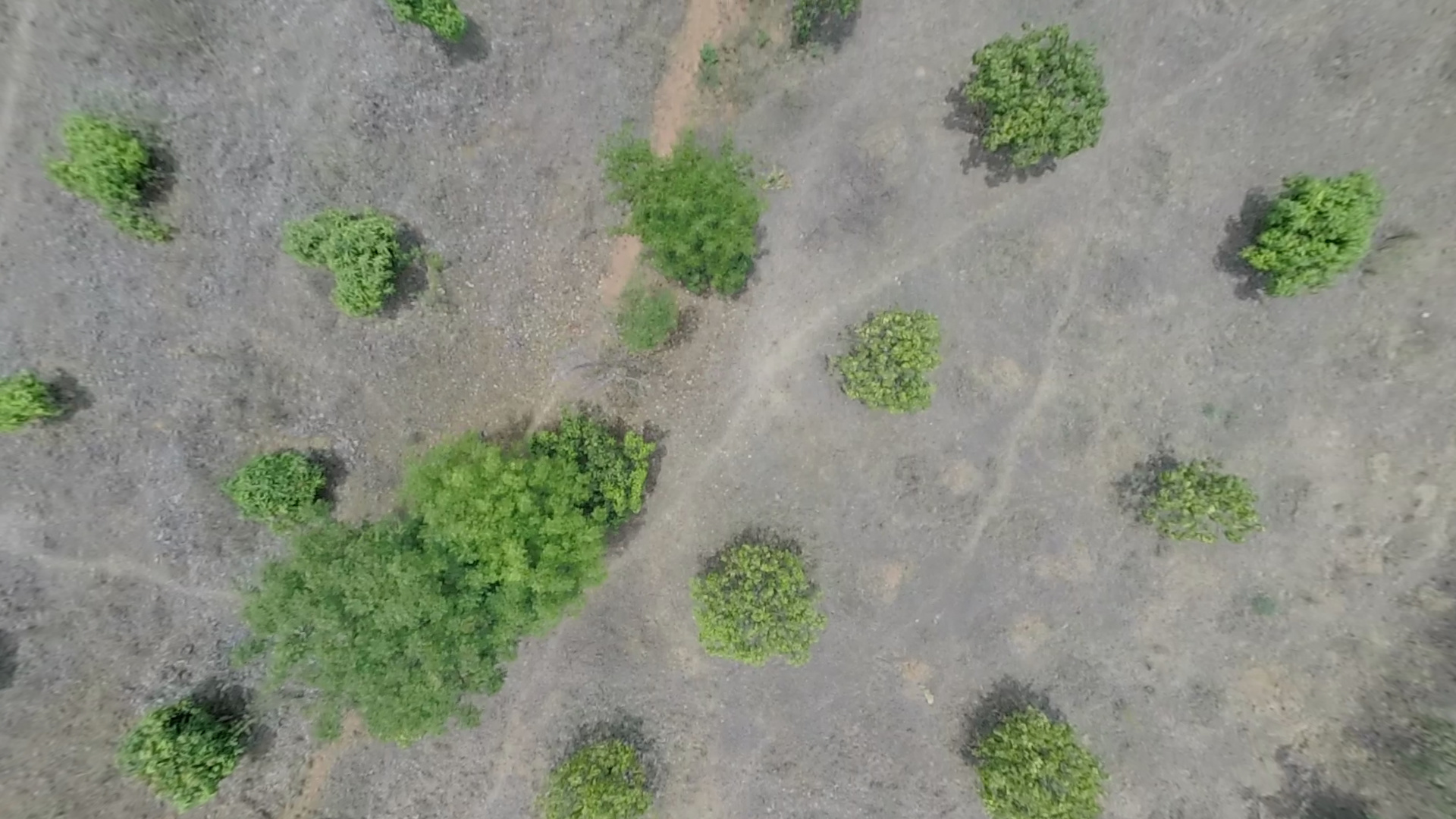} \hspace{1mm}\includegraphics[width=0.15\textwidth, height=0.15\textwidth]{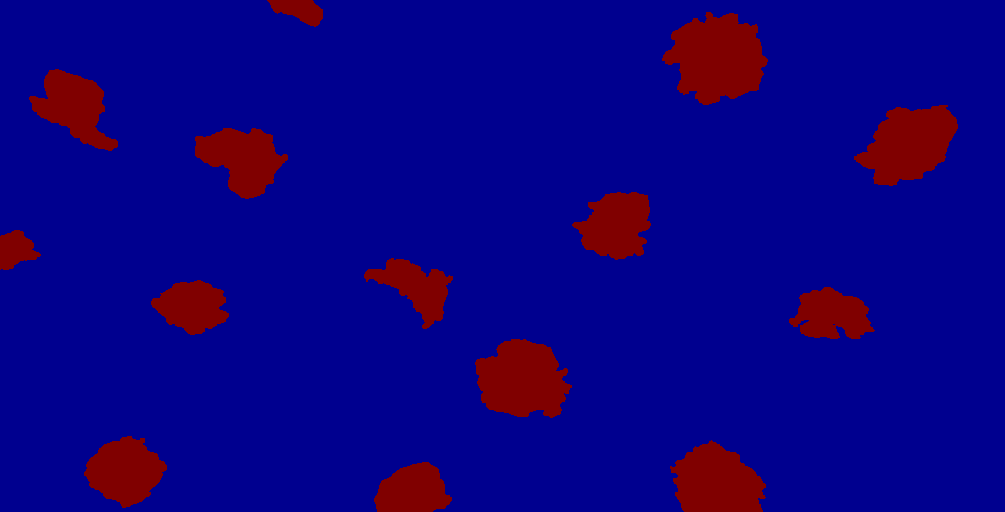} \hspace{1mm}\includegraphics[width=0.15\textwidth, height=0.15\textwidth]{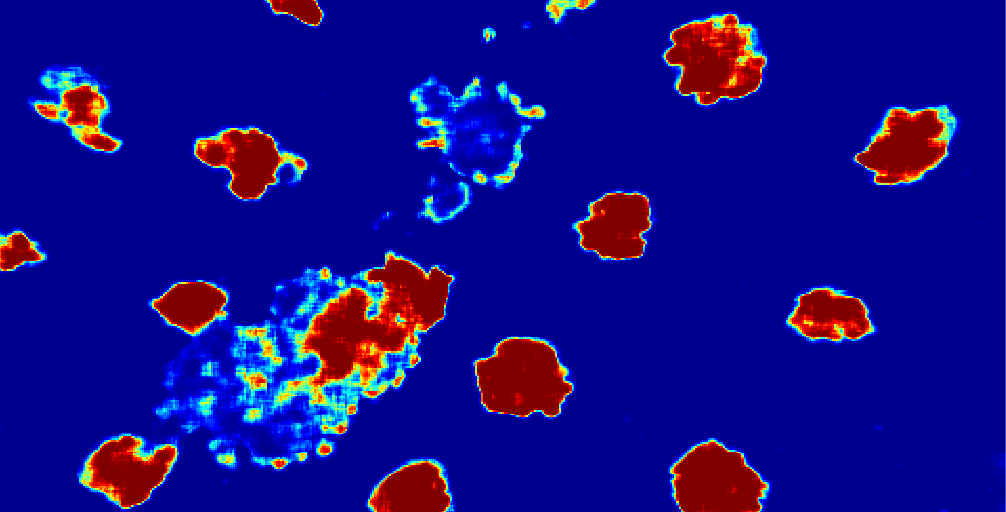} \hspace{1mm}\includegraphics[width=0.15\textwidth, height=0.15\textwidth]{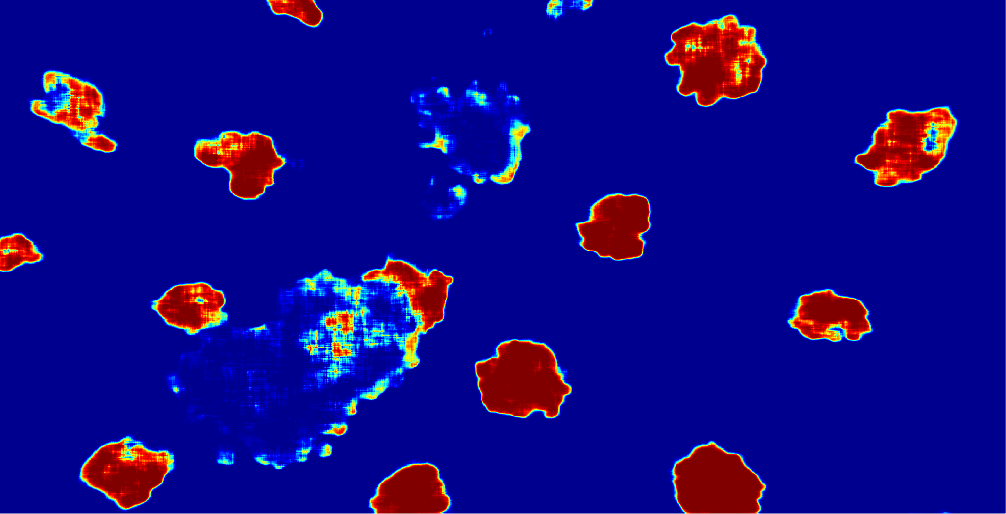} \hspace{1mm}\includegraphics[width=0.15\textwidth, height=0.15\textwidth]{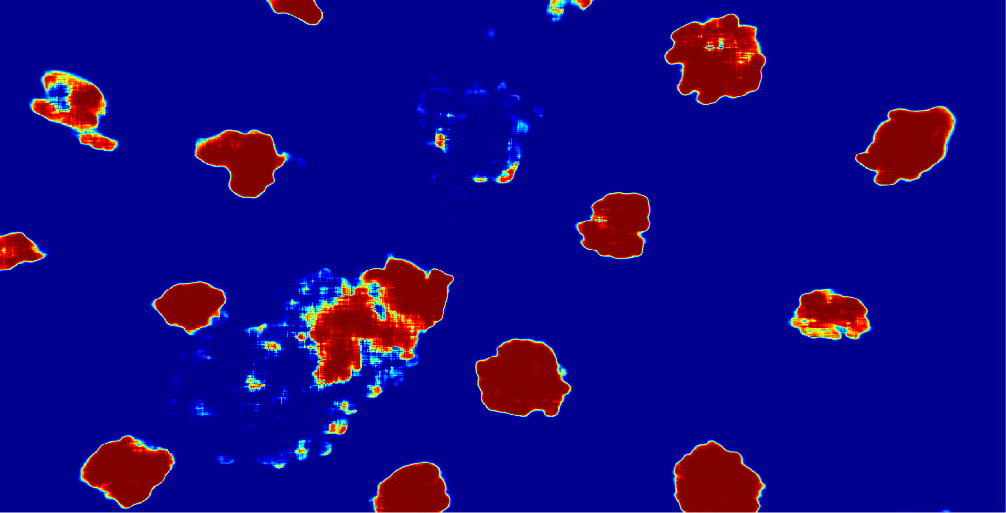} \hspace{1mm}\includegraphics[width=0.15\textwidth, height=0.15\textwidth]{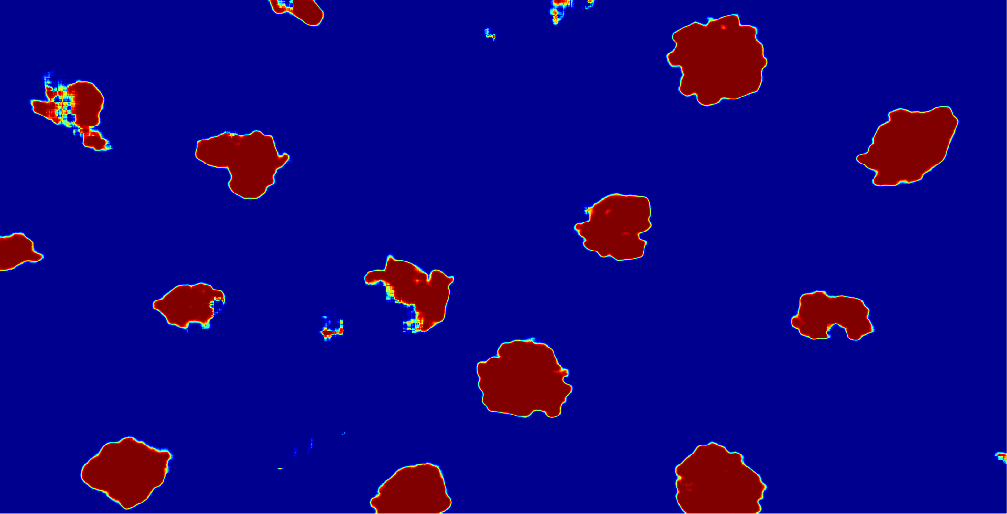} \vspace{1mm} \\
     (a)\hspace{0.13\textwidth}(b)\hspace{0.14\textwidth}(c)\hspace{0.14\textwidth}(d)\hspace{0.14\textwidth}(e)\hspace{0.13\textwidth}(f)
\end{minipage}
\caption{(a) Test images, (b) Heat maps of ground truth (c) Heatmaps of output from Arch-1 (d) Heatmaps of output from Arch-2 (e) Heatmaps of output from Arch-3 (f) Heatmaps of output from Mango Tree Net}
\label{fig:hm}  
\end{figure}

\begin{figure}
\centering
\begin{tabular}{cc}
\includegraphics[width=0.48\textwidth]{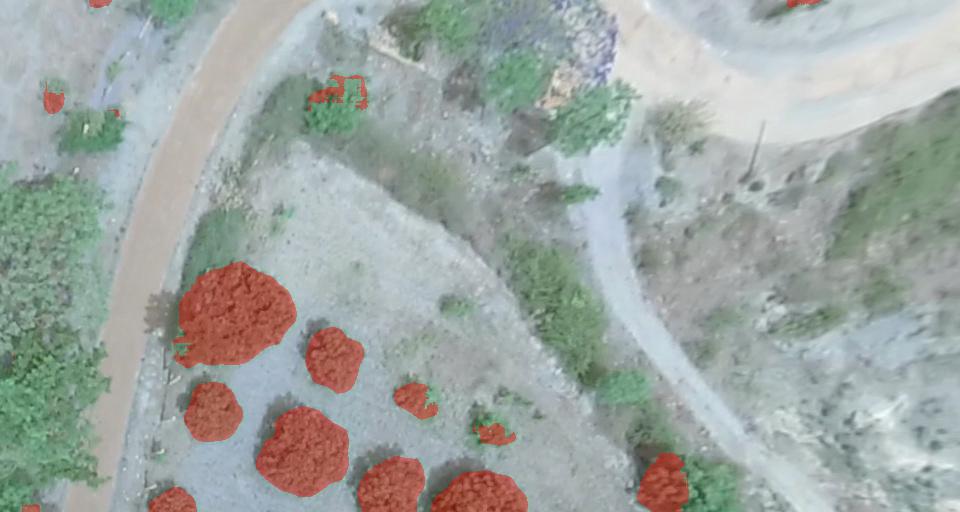} &
\includegraphics[width=0.48\textwidth]{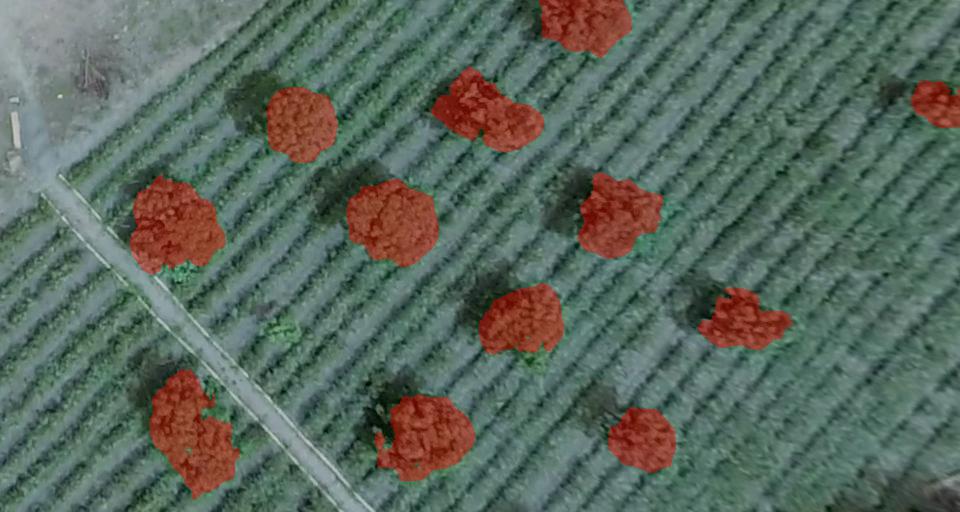} \\
(a) & (b) \\
\includegraphics[width=0.48\textwidth]{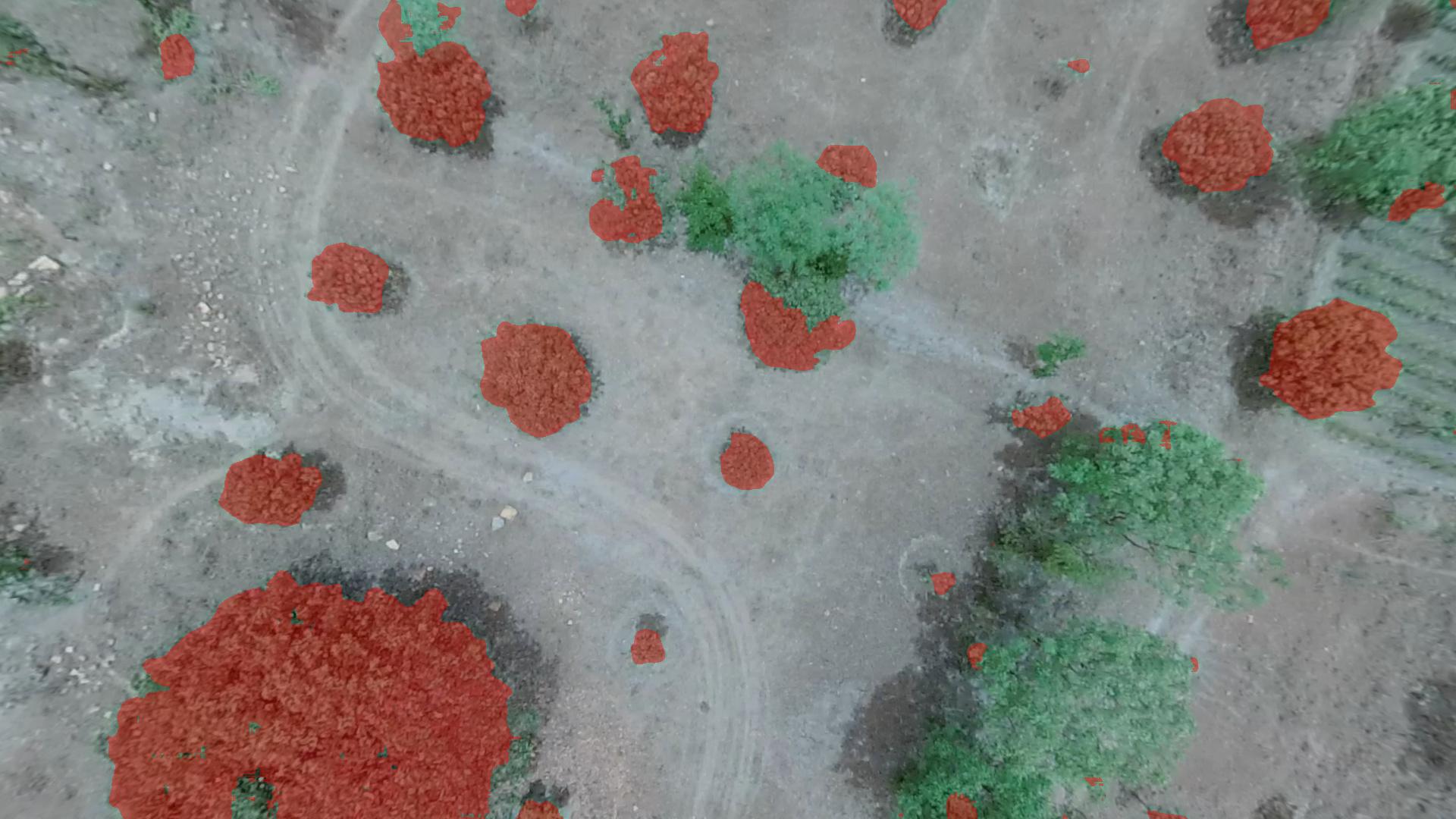} &
\includegraphics[width=0.48\textwidth]{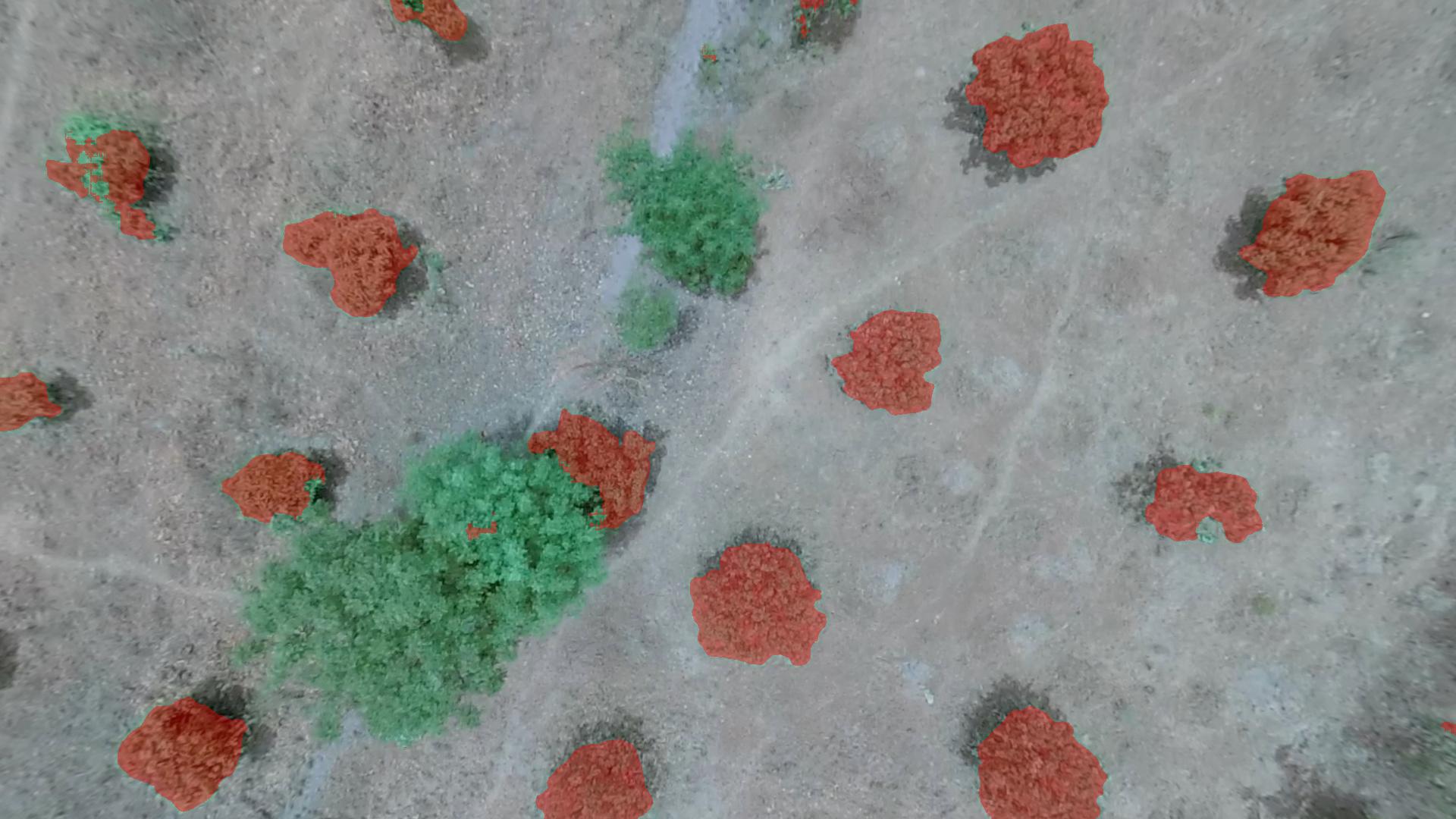} \\
(c) & (d) \\
\end{tabular}
\caption{Segmentation map overlaid on test images labelled Image $1$, $2$, $3$ and $4$ in (a), (b), (c) and (d) respectively.}
\label{fig:so}
\end{figure}

Four representative images opted for in-depth analysis are shown in figure \ref{fig:hm}(a) and the corresponding ground truths in figure \ref{fig:hm}(b). Figure \ref{fig:hm}(c)-(f) show heatmaps of outputs (prior threshold) from the architectures in consideration. Heatmaps presented indicate the probability ($0$-$1$) of each pixel belonging to mango tree class. The pixels in blue color have the least value ($0$) and those in red have the maximum value ($1$). It is clear from figure \ref{fig:hm} that Mango Tree Net assigns high probability to mango tree pixels and low probability to background pixels. Whereas, other architectures tend to assign relatively high probability to pixels of non-mango vegetation (part of background class). This leads to the increase in the number of false positives and consequent drop in performance. This indicates Mango Tree Net has sufficient learning capacity to distinguish mango trees from other vegetation. In figure \ref{fig:so}, the outputs of Mango Tree Net post thresholding are overlaid on test images for visualization.

In figure \ref{fig:so}(a)-(b) the stretch of shadow indicates that test dataset has images taken under different lighting conditions at different times of the day. Figure \ref{fig:so}(a) has been taken with drone tilting to change direction. The image has a variety of vegetation including trees, shrubs and stray/unwanted plants along with mango trees. In figure \ref{fig:so}(b), mango trees are planted between rows of tomato plants. The image was taken during descent of UAV. Figure \ref{fig:so}(c) has mango trees ranging from small young to large old trees. It shows trees partially occluded by non-mango vegetation. Figure \ref{fig:so}(d) has three non-mango trees with geometry and crown structure similar to that of a mango tree. The four representative images were chosen to demonstrate the diversity in test dataset and  challenges posed. Unlike traditional computer vision algorithms that exploit only low-level geometric and spectral properties, the Mango Tree Net learns complex high-level features. Despite variations in lighting conditions, shadowing, altitude and angle of image acquisition, and presence of similar vegetation, the Mango Tree Net can segment mango tree crowns correctly. This is evident from figure \ref{fig:so} and the results tabulated in tables \ref{tab:pe4} and \ref{tab:pec}.
	
The performance metrics for four representative images are tabulated in table \ref{tab:pe4}. It can be seen that Mango Tree Net fares well on precision, f1-score and accuracy metrics across all four images images. The heatmaps in figure \ref{fig:hm}(c) from Arch-1 assign high probability to pixels of background class containing non-mango vegetation. Post thresholding, those pixels are classified as mango tree class. Therefore, the count of false positives increases resulting in decrease of precision metric. The problem is ameliorated in Arch-2 and Arch-3 (figure \ref{fig:hm}(d) and (e)) and eliminated in Mango Tree Net (figure \ref{fig:hm}(f)). The same is reflected by the precision values in table \ref{tab:pe4}. The high recall values for Arch-1 across all images indicates least false negatives. This is because Arch-1 predicts most of vegetation as mango trees, as a result small number of mango tree pixels are misclassified. The performance of Mango Tree Net on recall metric is comparable to that of Arch-1. The Mango Tree Net also has highest values of f1-score and accuracy for all images. It is consistent across all metrics of performance. The comparable values of accuracy across all architectures is attributed to class imbalance.

The performance metrics in table \ref{tab:pec} follow a similar trend as that of representative images. The values are obtained using total number of true positives ($TP$), false positives ($FP$) and false negatives ($FN$) across 36 images in the test dataset. It can be concluded from the results that Mango Tree Net is the superior among the four and suitable architecture for semantic segmentation of mango tree crowns.

\begin{table}[]
\centering
\caption{Detailed performance evaluation on semantic segmentation task using four representative images}
  \begin{tabular}{cccccc}\toprule
    \textbf{Test image} & \textbf{Architecture} & \textbf{Precision} & \textbf{Recall} & \textbf{F1-Score} & \textbf{Accuracy}\\  \midrule
	\multirow{4}{*}{Image 1} & Arch-1 & 0.4332 & \textbf{0.9876} & 0.6023 & 0.9278 \\
	                         & Arch-2 & 0.6877 & 0.8981 & 0.7789 & 0.9718 \\
							 & Arch-3 & 0.6749 & 0.9412 & 0.7861 & 0.9717 \\
							 & Mango Tree Net & \textbf{0.7645} & 0.9433 & \textbf{0.8445} & \textbf{0.9808} \\ \midrule
    \multirow{4}{*}{Image 2} & Arch-1 & 0.7203 & \textbf{0.8885} & 0.7956 & 0.9408 \\
	                         & Arch-2 & 0.7472 & 0.8361 & 0.7892 & 0.9420 \\
							 & Arch-3 & 0.8174 & 0.8924 & 0.8533 & 0.9602 \\
							 & Mango Tree Net & \textbf{0.9553} & 0.8640 & \textbf{0.9074} & \textbf{0.9771} \\ \midrule
	\multirow{4}{*}{Image 3} & Arch-1 & 0.8049 & \textbf{0.9357} & 0.8654 & 0.9598 \\
	                         & Arch-2 & 0.9052 & 0.8806 & 0.8927 & 0.9708 \\
							 & Arch-3 & 0.8891 & 0.8668 & 0.8778 & 0.9667 \\
							 & Mango Tree Net & \textbf{0.9401} & 0.9113 & \textbf{0.9255} & \textbf{0.9798} \\ \midrule
    \multirow{4}{*}{Image 4} & Arch-1 & 0.7541 & \textbf{0.9325} & 0.8339 & 0.9630 \\
	                         & Arch-2 & 0.9084 & 0.9190 & 0.9137 & 0.9827 \\
							 & Arch-3 & 0.8298 & 0.9363 & 0.8799 & 0.9745 \\
							 & Mango Tree Net & \textbf{0.9610} & 0.9162 & \textbf{0.9381} & \textbf{0.9880} \\ \bottomrule							 						
	\end{tabular}
\label{tab:pe4}
\end{table}

\begin{table}[]
\centering
\caption{Performance evaluation on semantic segmentation task across entire test dataset}
  \begin{tabular}{ccccc}\toprule
    \textbf{Architecture} & \textbf{Precision} & \textbf{Recall} & \textbf{F1-Score} & \textbf{Accuracy}\\ \midrule
		Arch-1 & 0.7494 & \textbf{0.9173} & 0.8249 & 0.9474 \\
	    Arch-2 & 0.8747 & 0.8822 & 0.8784 & 0.9670 \\
		Arch-3 & 0.8479 & 0.8669 & 0.8573 & 0.9610 \\
		Mango Tree Net & \textbf{0.9085} & 0.8941 & \textbf{0.9012} & \textbf{0.9735} \\ \bottomrule
  \end{tabular}
\label{tab:pec}
\end{table}

\begin{figure}[]
\centering
  \begin{minipage}{0.49\linewidth}
  \centering
    \def\svgwidth{0.99\columnwidth}
    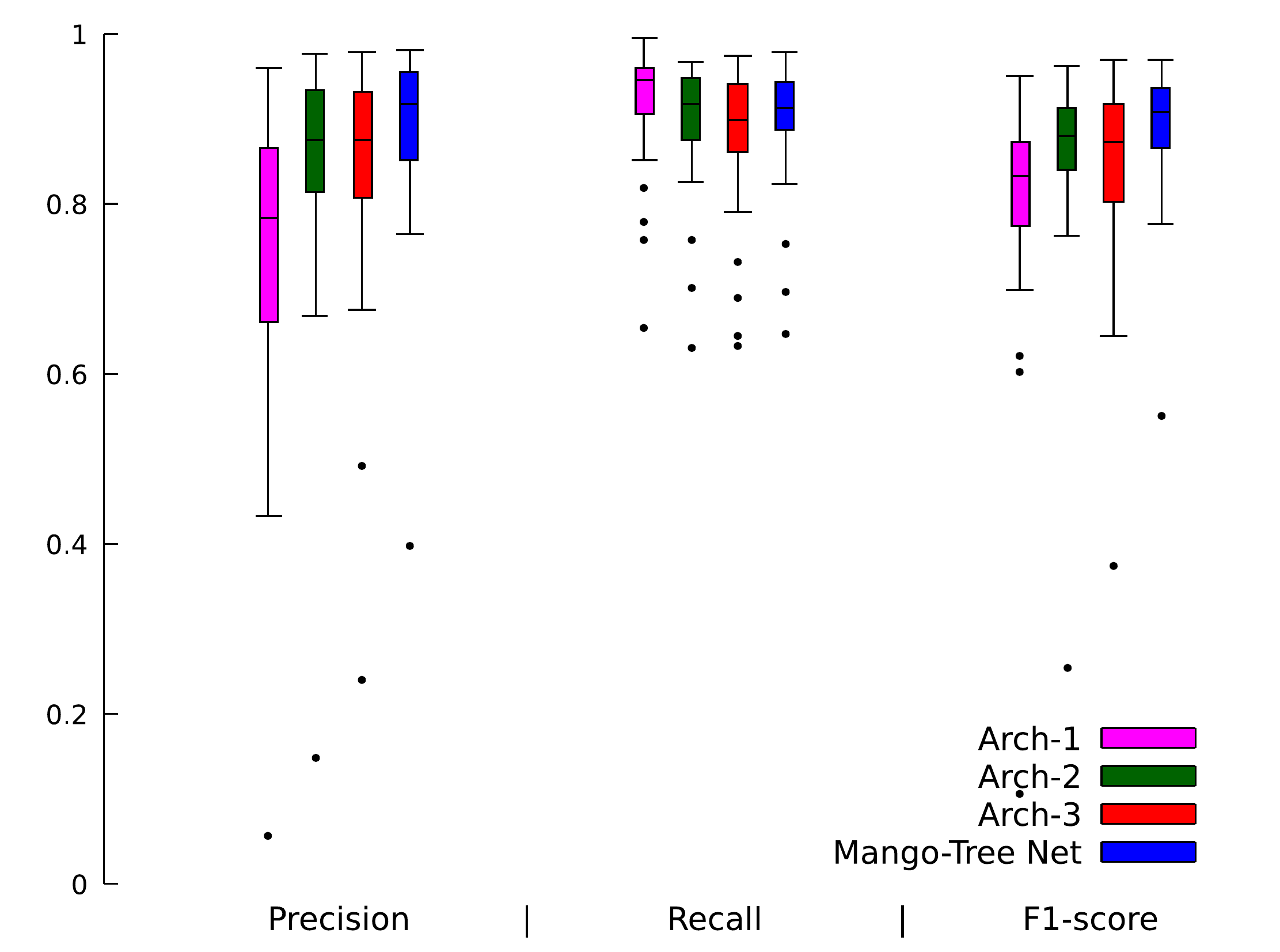
    \caption{Visualization of performance on semantic segmentation task using boxplots of precision, recall and f1-score metrics}
      \label{fig:bp}
  \end{minipage}
  \begin{minipage}{0.49\linewidth}
  \centering
    \def\svgwidth{0.96\columnwidth}
    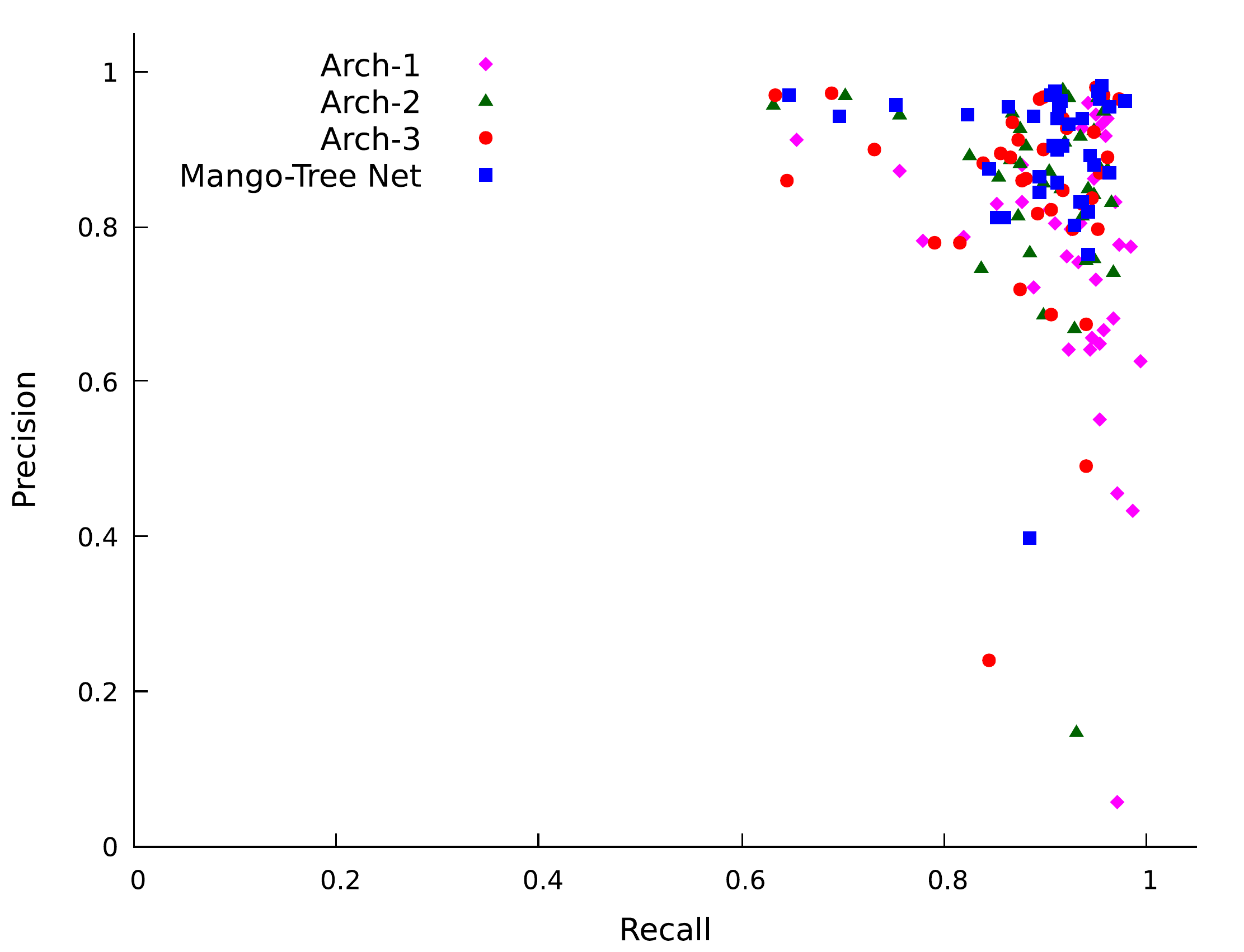
  \caption{Visualization of performance on semantic segmentation task using scatter plot of precision v/s recall metrics}
    \label{fig:pr}
  \end{minipage}   
\end{figure}

Figure \ref{fig:bp} provides the visualization of performance metrics precision, recall and f1-score for 36 images as box plots. The range and interquartile range are small across metrics for Mango Tree Net compared to other architectures. This indicates consistency of its performance across images. The median values of precision and f1-score of Mango Tree Net lie above the median values of other architectures. The visualization of precision-recall scatter plot is provided in figure \ref{fig:pr}. The points of Mango Tree Net for a given image lie relatively closer to (1, 1) which is more desirable.

\subsection{Individual crown detection results and visualization}
\label{sbsec:icd}
As mentioned earlier, four images with large number of trees appearing to be touching/overlapping are chosen for performance evaluation of individual crown detection. The detection through segmentation method discussed in section \ref{sec:testing} is used. The images chosen have been taken with UAV tilted at various angles which makes large number of closely planted trees to appear to be overlapping, and hence, more suitable for performance evaluation. The total number of mango trees in ground truth are determined by manually counting during annotation. There are $55$ mango trees in image $1$, $56$ in image $2$, $79$ in image $3$ and $107$ in image $4$. There are $297$ mango trees in total.

\begin{table}[h]
\centering
\caption{Performance evaluation on individual crown detection task on test dataset}
  \begin{tabular}{cccc}\toprule
    \textbf{Test Image} & \textbf{Precision} & \textbf{Recall} & \textbf{F1-Score}\\ \midrule
		Image 1 & 0.9804 & 0.9091 & 0.9434 \\
	    Image 2 & 0.9600 & 0.8571 & 0.9056 \\
		Image 3 & 0.9692 & 0.7975 & 0.8750 \\
		Image 4 & 1.0000 & 0.8785 & 0.9353 \\ \bottomrule
  \end{tabular}
\label{tab:peic}
\end{table}

\begin{table}[h]
\centering
\caption{Comparison of individual crown detection with 2-class and 3-class Mango Tree Net}
  \begin{tabular}{cccc}\toprule
    \textbf{Model} & \textbf{Precision} & \textbf{Recall} & \textbf{F1-Score}\\ \midrule
		2-class MTN & 0.9573 & 0.6801 & 0.7953 \\
	    3-class MTN & \textbf{0.9808} & \textbf{0.8586} & \textbf{0.9156} \\ \bottomrule
  \end{tabular}
\label{tab:peict}
\end{table}

\begin{figure}[]
\centering
\includegraphics[width=0.95\textwidth]{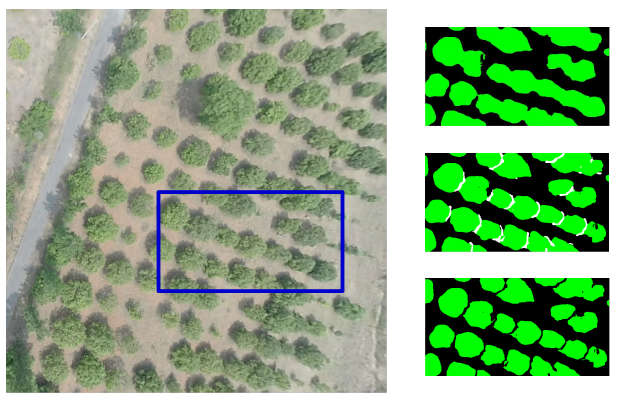}
\caption{The blue rectangle highlights an area containing trees that appear to overlap/touch. The three images on the right are semantic segmentation outputs of highlighted area from 2-class and 3-class Mango Tree Net. Top right image is from 2-class model. Middle right image is from 3-class model before setting boundary pixels as background. Bottom right is also from 3-class model before setting boundary pixels as background }
\label{fig:itcd1}
\end{figure}

\begin{figure}[ht!]
\centering
\begin{tabular}{ccc}
\includegraphics[width=0.3\textwidth]{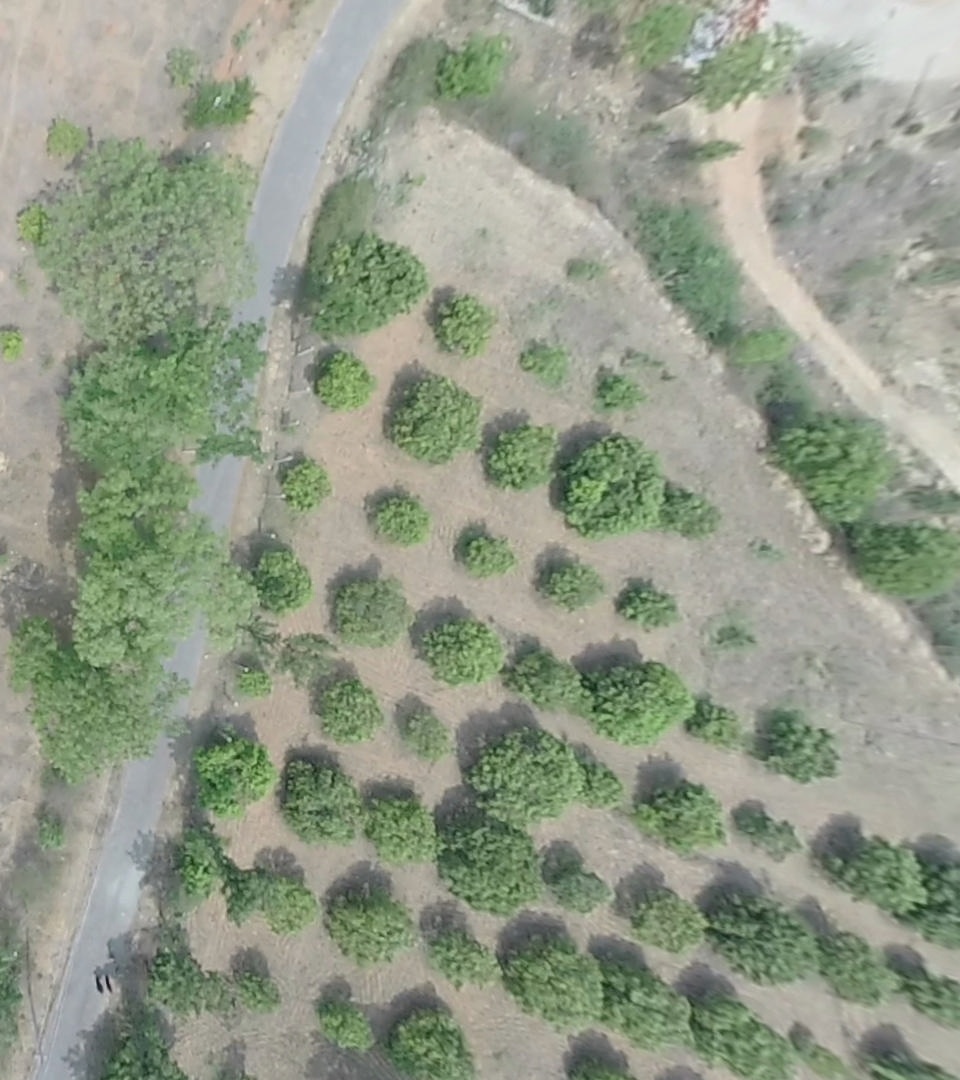} & 
\includegraphics[width=0.3\textwidth]{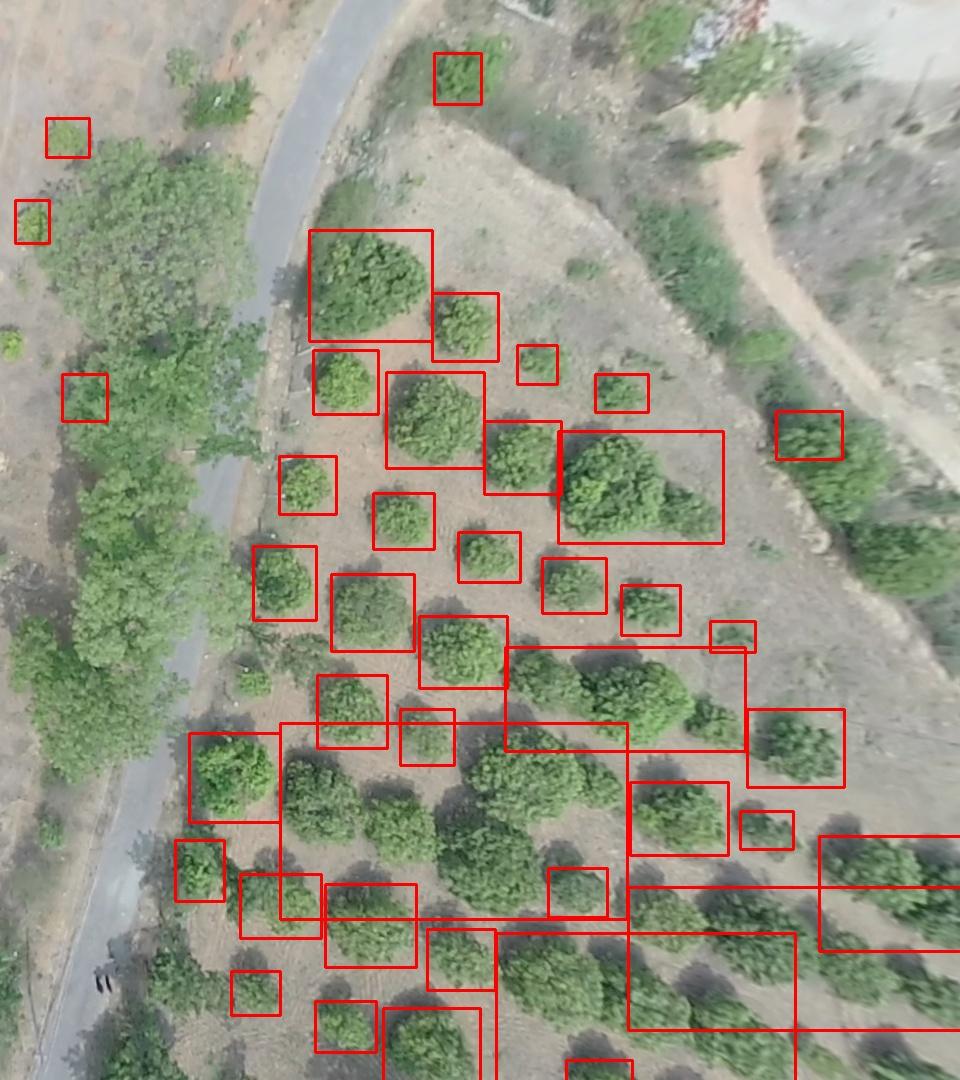} & 
\includegraphics[width=0.3\textwidth]{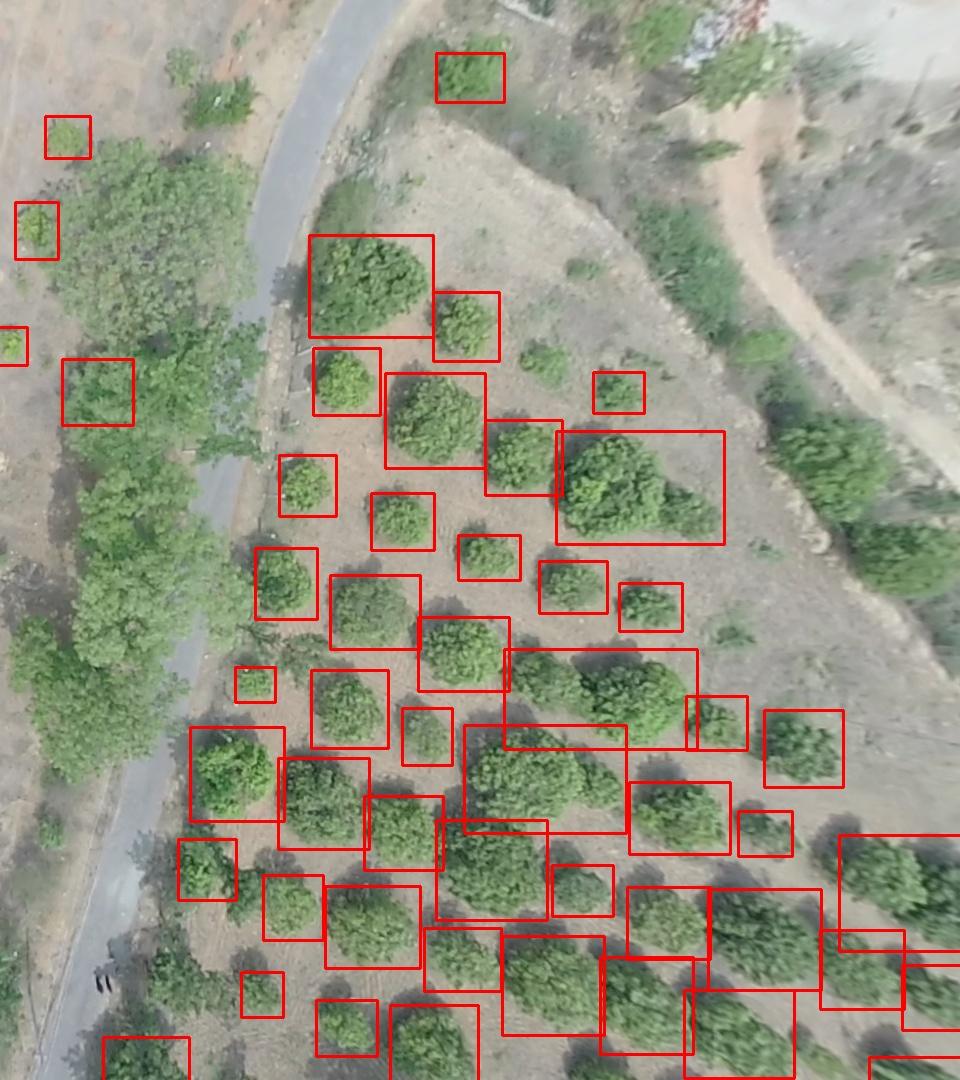} \\
\includegraphics[width=0.3\textwidth]{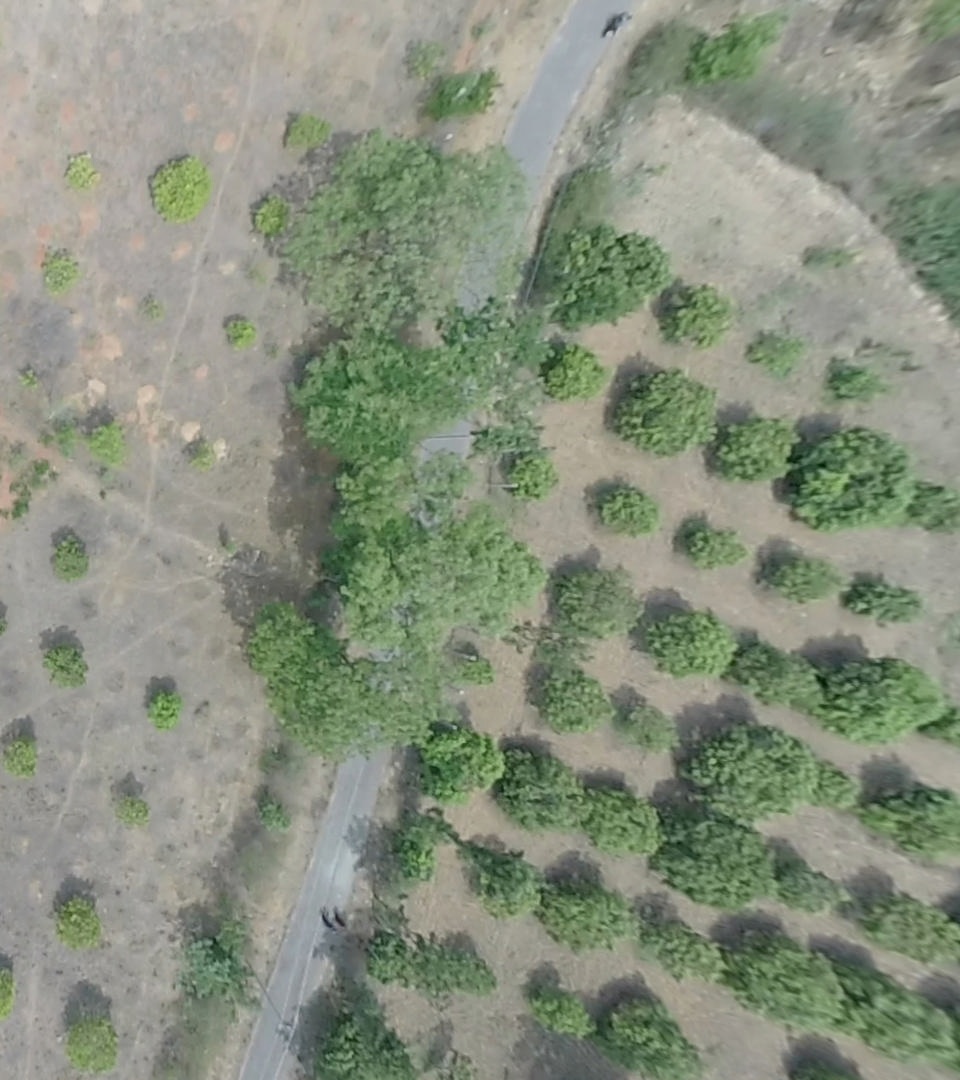} & 
\includegraphics[width=0.3\textwidth]{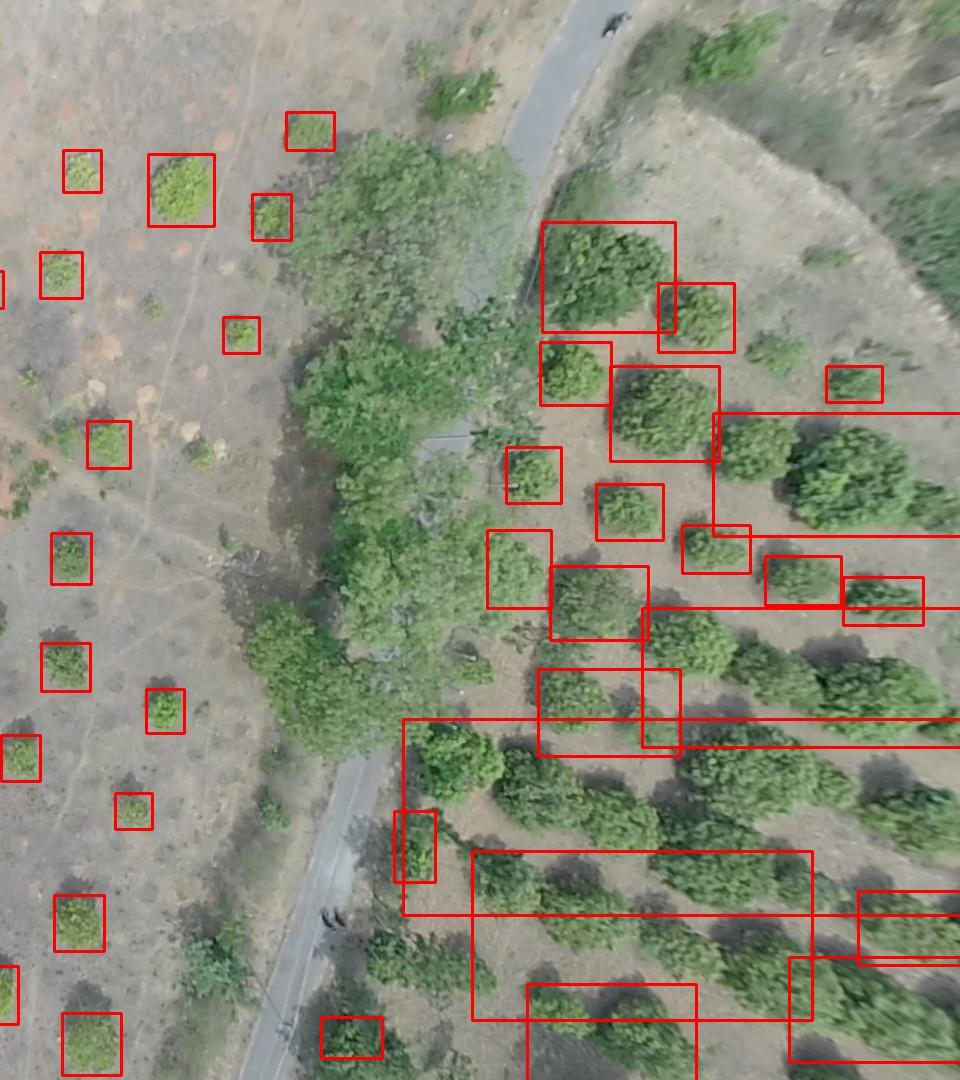} & 
\includegraphics[width=0.3\textwidth]{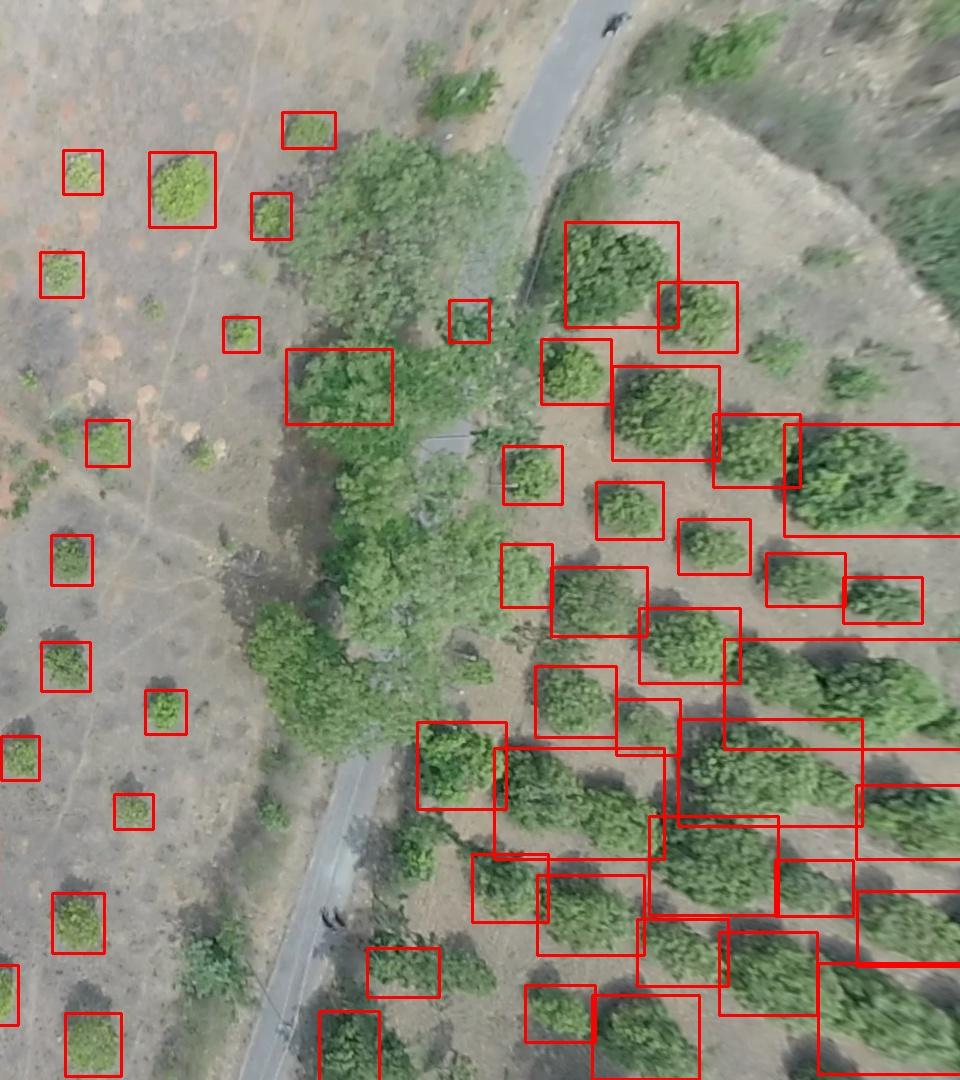} \\
\includegraphics[width=0.3\textwidth]{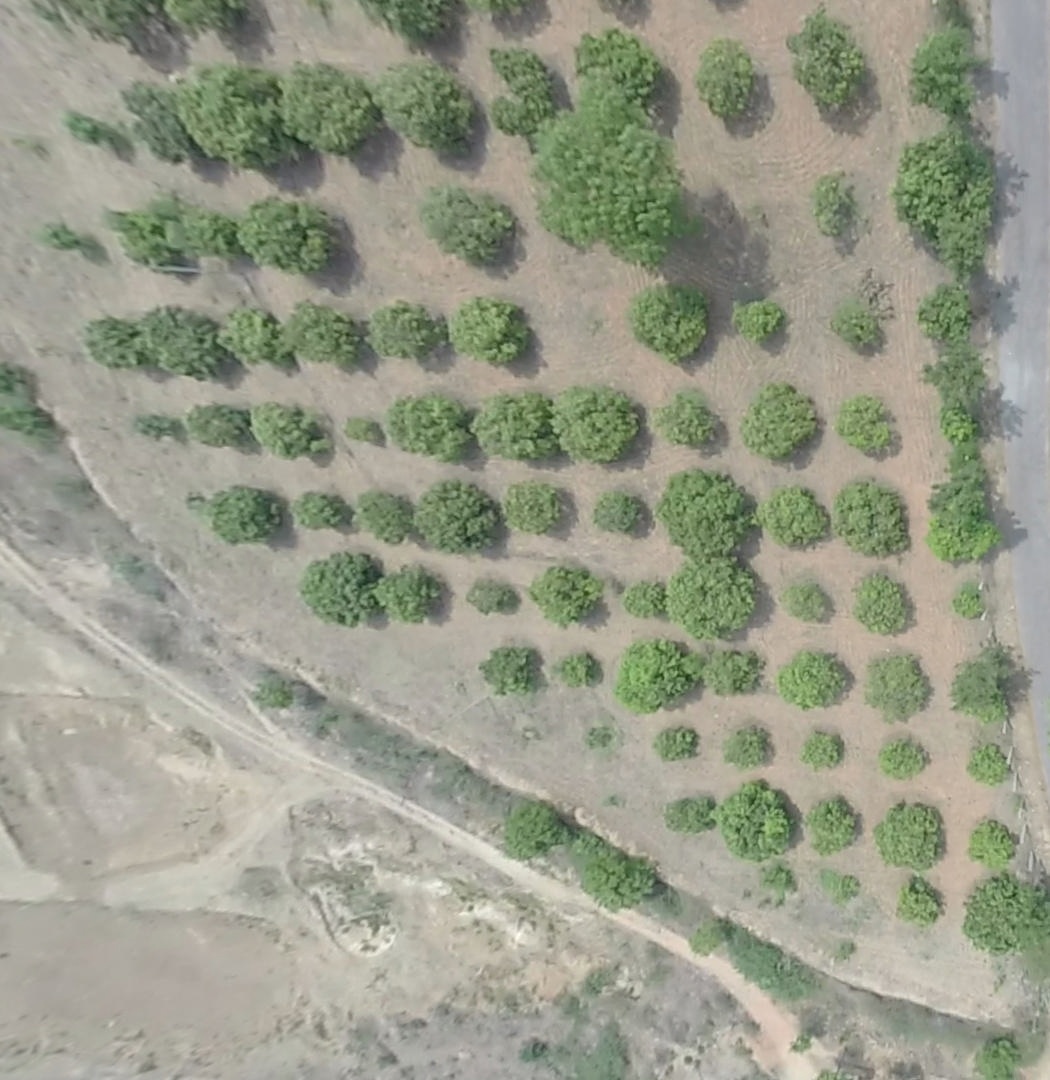} & 
\includegraphics[width=0.3\textwidth]{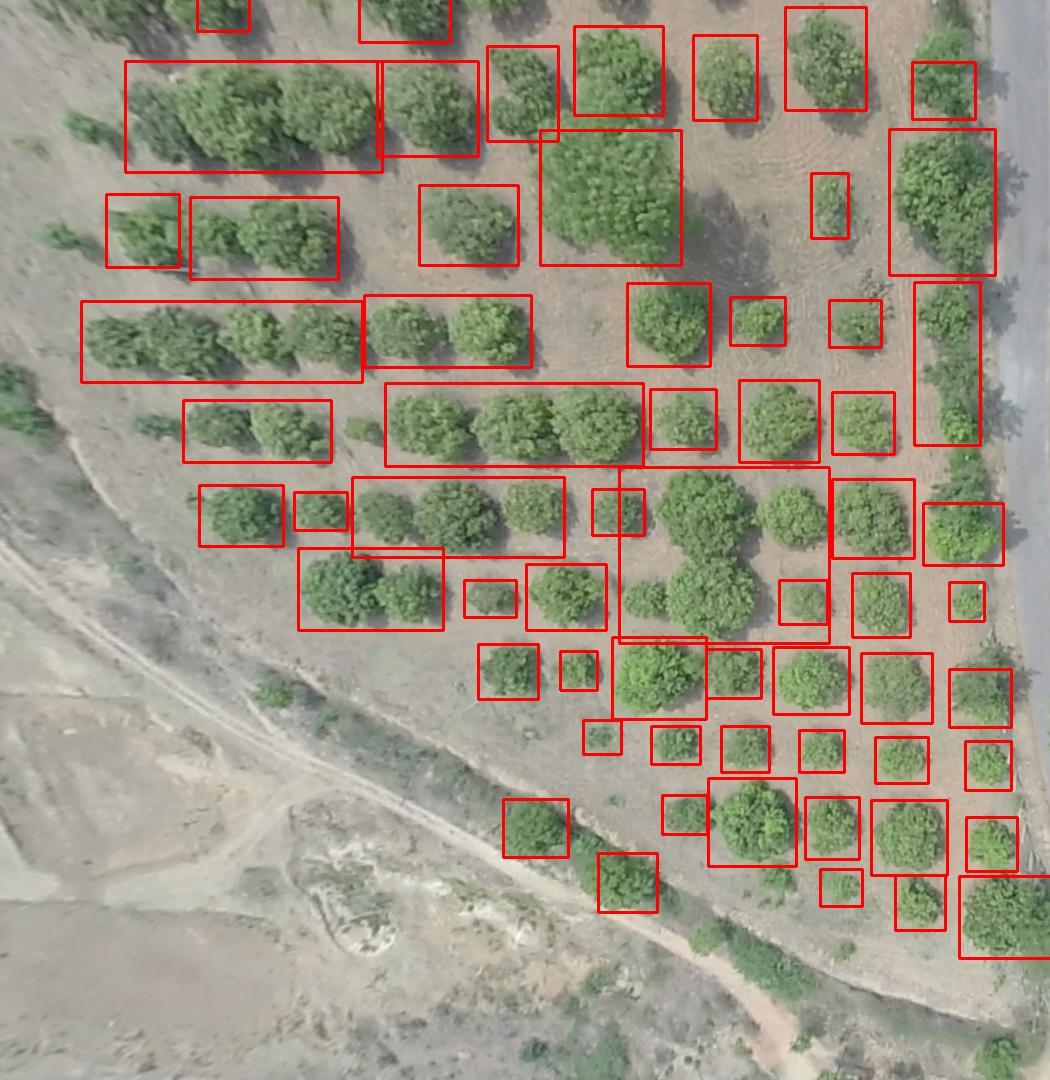} & 
\includegraphics[width=0.3\textwidth]{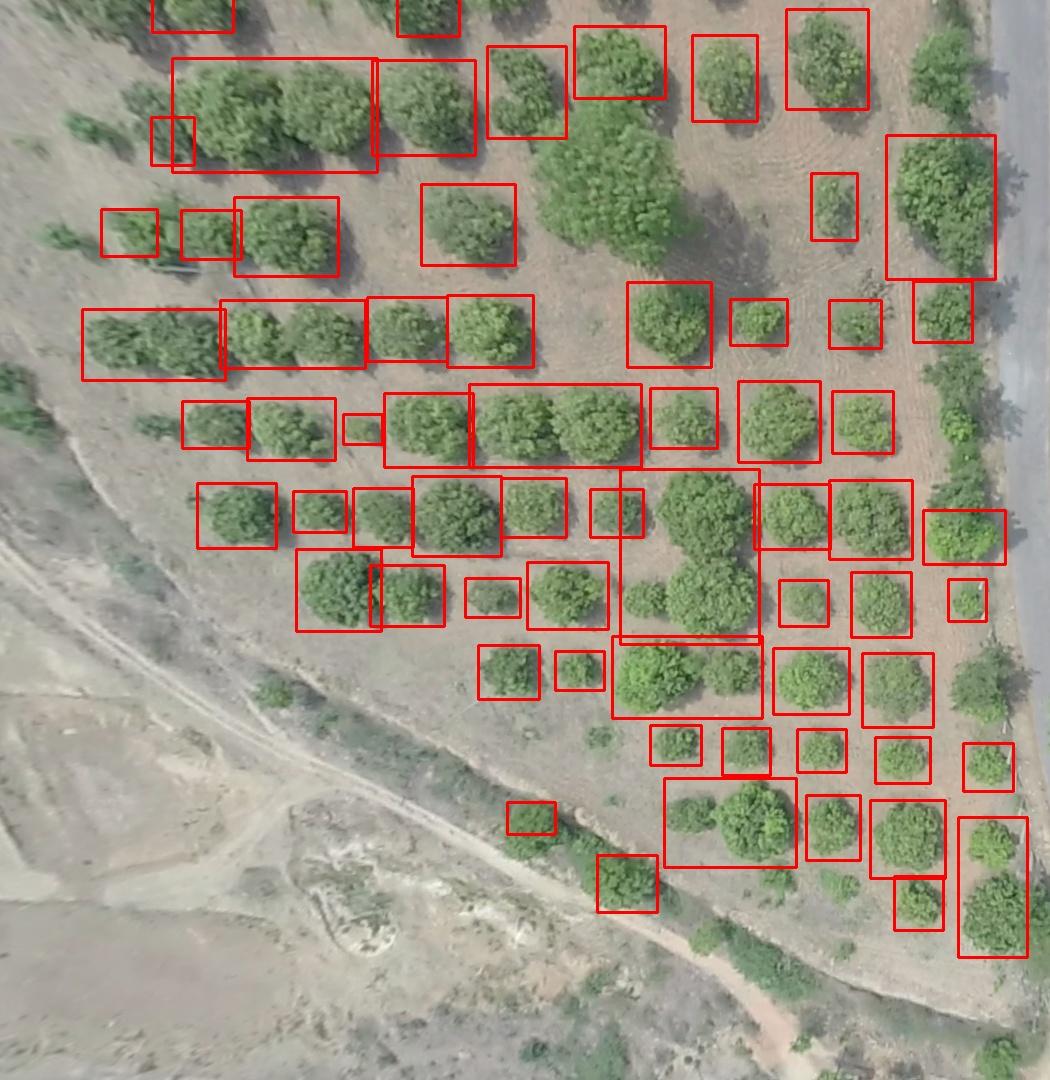} \\
\includegraphics[width=0.3\textwidth]{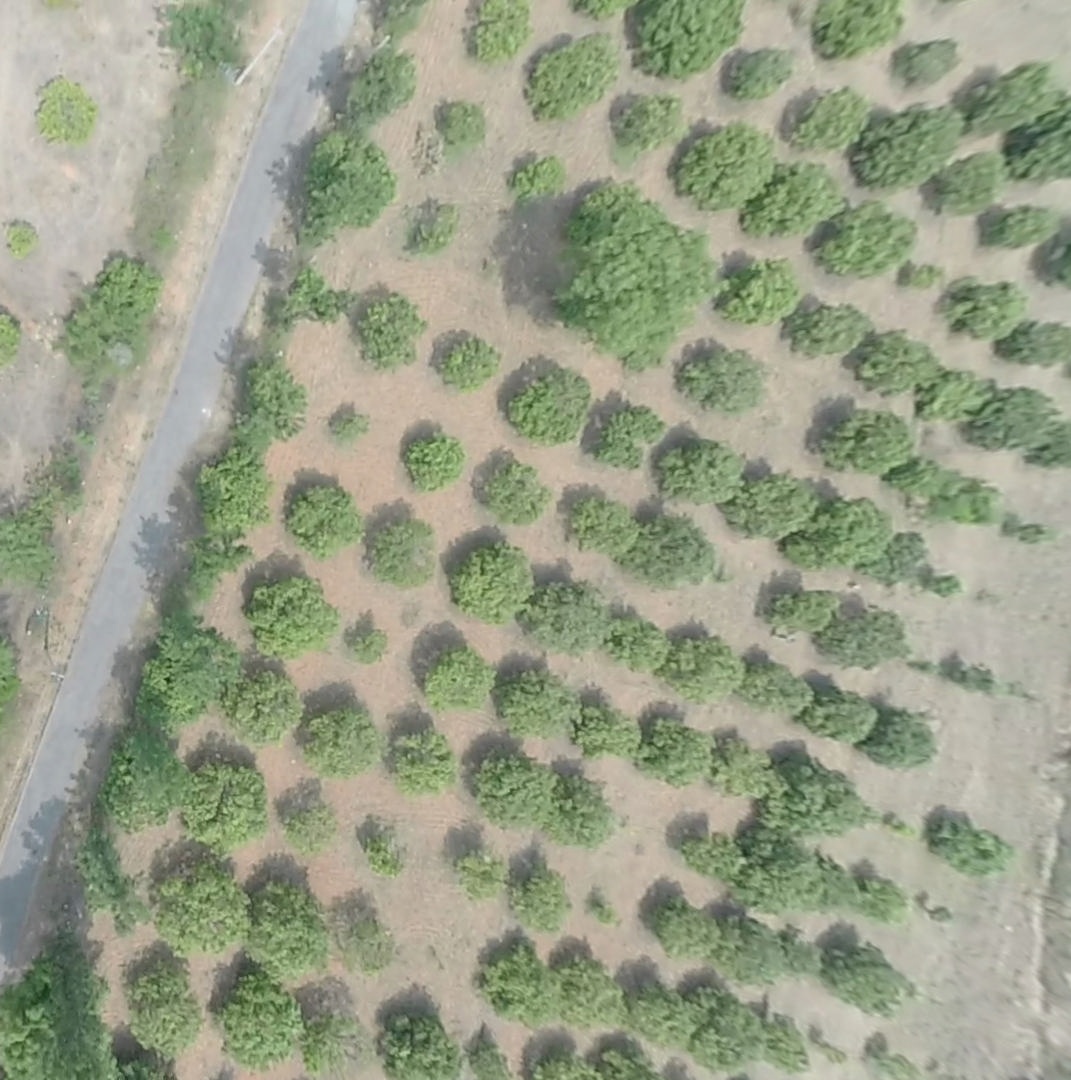} & 
\includegraphics[width=0.3\textwidth]{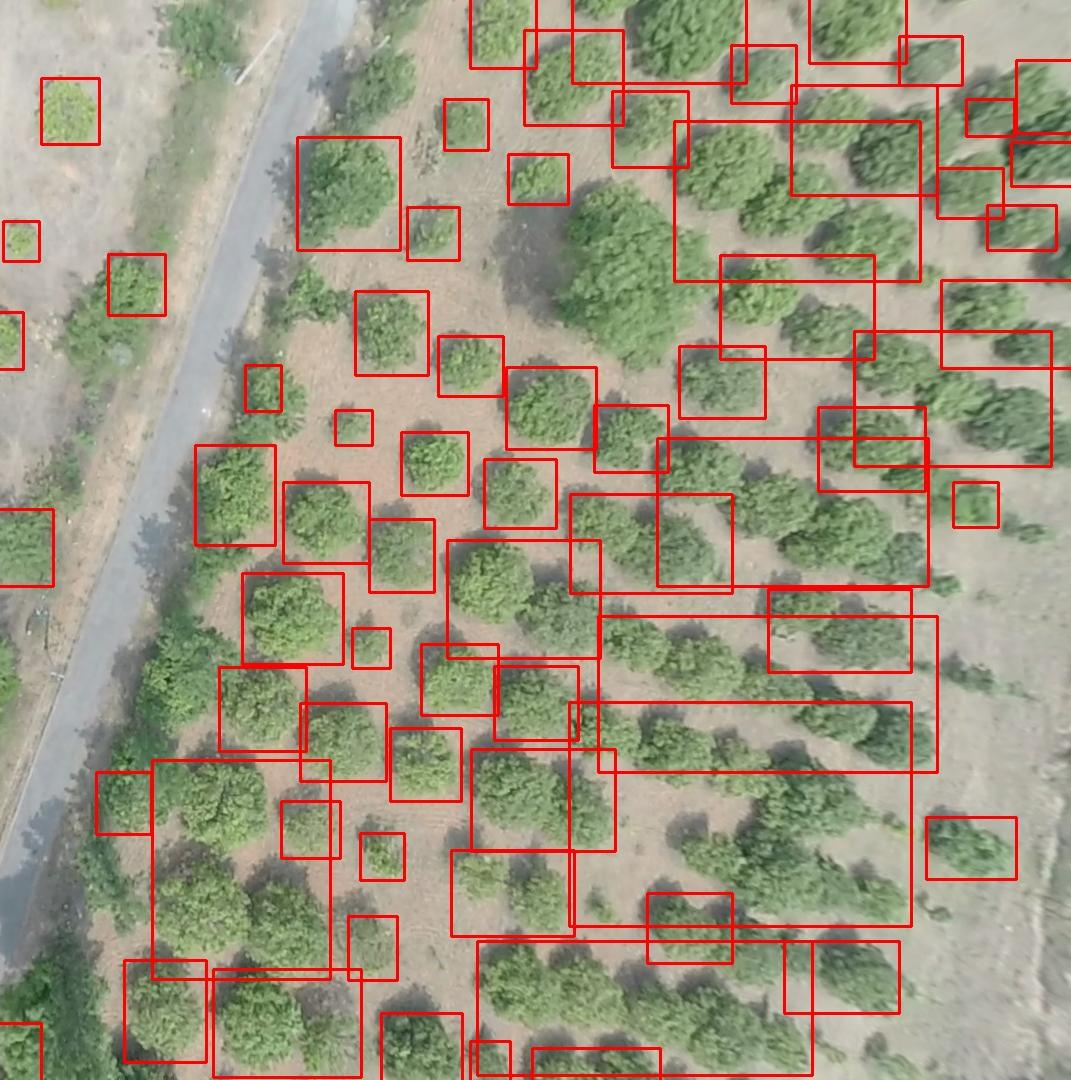} & 
\includegraphics[width=0.3\textwidth]{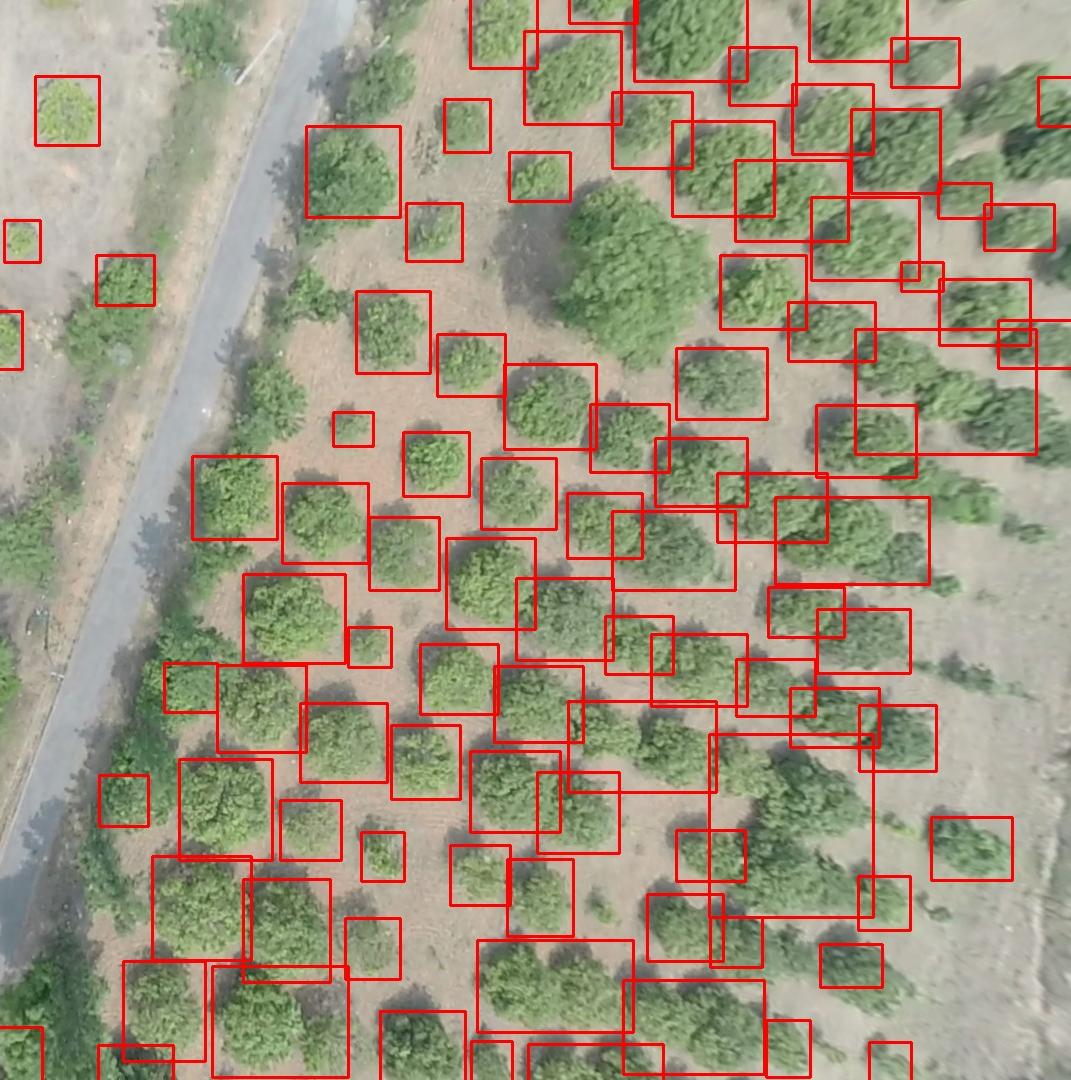} \\
(a) & (b) & (c)
\end{tabular}
\caption{(a) Test Images labelled as Image $1$, $2$, $3$ and $4$ repectively (b) Trees detected using 2-class model (c) Trees detected using 3-class model}
\label{fig:itcd2}
\end{figure}

In figure \ref{fig:itcd1}, a region with touching/overlapping tree crowns is highlighted with blue box. The segmentation output of highlighted region from 2-class Mango Tree Net is on the top right. The segmentation output of highlighted region from 3-class Mango Tree Net, after setting boundary class pixels as background class, is on the bottom right. The bottom right image has touching/overlapping trees separated in semantic segmented output. Thus, performing connected object detection on bottom right image yields much better detection accuracy. The individual crown detections obtained using 2-class and 3-class models along with the test image are shown in figure \ref{fig:itcd2}. The detections obtained using 2-class model are used to demonstrate the effectiveness of 3-class model. A sharp increase in the count of detections was obtained using 3-class model.

The results of individual crown detection through segmentation using 3-class model for each image are tabulated in table \ref{tab:peic}. Table \ref{tab:peict} compares the results of detection through segmentation using 2-class and 3-class Mango Tree Nets. The results in table \ref{tab:peict} are calculated using sum total of true positives ($TP$), false positives ($FP$) and false negatives ($FN$) across $4$ images. The improvement in the count of number of trees with 3-class model is evident. The precision is high using both models which indicates small number of false positives in both cases. The value of recall is $0.6801$ using 2-class model. This implies that the count obtained is $68.01$\% of actual number. The value of recall is $0.8586$ using 3-class model for counting. Thus, the count of number of trees obtained using 3-class model is $85.86$\% of the actual number. The f1-score is $0.9156$ for 3-class model opposed to $0.7953$ for 2-class model.

\section{Conclusion}
\label{sec:conclusion}

This paper successfully accomplishes the task of semantic segmentation and detection of individual mango tree crowns from high resolution UAV imagery. The supervised learning approach using Mango Tree Net, an FCN based model, is developed. The Mango Tree Net was trained to produce 2-class and 3-class segmentation maps. For the semantic segmentation task, the 2-class segmentation model was used that classifies pixels into mango tree and background classes. For individual mango tree crown detection task, the 3-class segmentation model was used that classifies pixels into mango tree, boundary separating touching/overlapping crowns and background classes. Individual mango tree crowns are detected by detecting connected objects in segmentation output of 3-class model. The proposed method for detecting individual tree crowns is novel and no past work was found through literature survey that used the method for said task. 

The proposed methods for segmentation and detection are robust and work despite variations in multitude of factors such as scale, occlusion, lighting conditions and surrounding vegetation. This is demonstrated by the results. The Mango Tree Net architecture is light and hence, suitable for real-time applications. Though the experimental site considered for this work is an orchard, the methods can further be extended to detect homogeneous, dense or isolated trees in forests.

\clearpage

\bibliography{references}
\bibliographystyle{abbrvnat}

\end{document}